\documentclass[sn-mathphys-num]{sn-jnl}

\usepackage{graphicx}%
\usepackage{multirow}%
\usepackage{amsmath,amssymb,amsfonts}%
\usepackage{amsthm}%
\usepackage{mathrsfs}%
\usepackage[title]{appendix}%
\usepackage{textcomp}%
\usepackage{manyfoot}%
\usepackage{booktabs}%
\usepackage{algorithm}%
\usepackage{algorithmicx}%
\usepackage{algpseudocode}%
\usepackage{listings}%

\theoremstyle{thmstyleone}%
\theoremstyle{thmstyletwo}%

\theoremstyle{thmstylethree}%
\newtheorem{definition}{Definition}%

\usepackage[table]{xcolor}
\usepackage{amssymb}
\usepackage{lipsum}
\usepackage{lscape}
\usepackage{graphicx}%
\usepackage{multirow}%
\usepackage{amsmath,amssymb,amsfonts}%
\usepackage{amsthm}%
\usepackage{amsfonts}
\usepackage{mathrsfs}%
\usepackage[title]{appendix}%
\usepackage{textcomp}%
\usepackage{manyfoot}%
\usepackage{booktabs}%
\usepackage{algorithm}%
\usepackage{algorithmicx}%
\usepackage{algpseudocode}%
\usepackage{listings}%
\usepackage{url}
\usepackage{physics}
\usepackage{amsmath}
\usepackage{tikz}
\usepackage[utf8]{inputenc}
\usepackage{pgfplots}
\DeclareUnicodeCharacter{2212}{−}
\usepgfplotslibrary{groupplots,dateplot}
\usetikzlibrary{patterns,shapes.arrows}
\pgfplotsset{compat=newest}
\usepackage{mathdots}
\usepackage{yhmath}
\usepackage{cancel}
\usepackage{color}
\usepackage{siunitx}
\usepackage{array}
\usepackage{multirow}
\usepackage{amssymb}
\usepackage{gensymb}
\usepackage{tabularx}
\usepackage{extarrows}
\usepackage{booktabs}
\usetikzlibrary{fadings}
\usetikzlibrary{patterns}
\usetikzlibrary{shadows.blur}
\usetikzlibrary{shapes}
\usepackage{listings}
\usepackage{graphicx} 
\usepackage{fancybox}
\usepackage{adjustbox}
\usepackage{twoopt} 
\usepackage{subcaption}
\usepackage{amsfonts} 
\usepackage{amssymb}
\usepackage{booktabs, multirow} 
\usepackage{soul}
\usepackage{changepage,threeparttable} 
\usepackage{forest}
\usepackage{pdflscape}
\usepackage{adjustbox}
\usetikzlibrary{shadows, trees, positioning}
\usepackage{booktabs}
\usepackage{xltabular}
\usepackage{tabularray}
\usepackage{enumerate}
\usetikzlibrary{shapes,arrows,positioning}
\usepackage[many]{tcolorbox}
\usetikzlibrary{calc}
\tcbuselibrary{skins}
\usepackage{dashrule}
\usetikzlibrary{decorations.pathreplacing}
\usepackage{titlesec}
\usepackage{float}

\newcommand{\newtext}[1]{\textcolor{black}{#1}}

\newcommand{\newtexttwo}[1]{\textcolor{black}{#1}}

\newcommand\marklessfootnote[1]{
    \addtocounter{footnote}{1} 
    \footnotetext{#1}
}
\renewcommand{\vec}[1]{\mathbf{#1}}
\newcommandtwoopt{\insertboxtwo}[3][1.0][4.0cm]{\noindent\fbox{\begin{minipage}[c]{#1\textwidth}\parbox[c][#2]{#1\textwidth}{#3}\hfill\end{minipage}}}
\newcommand{\casestudybox}[2]{
\newtext{\textbf{\noindent Scenario: #1}\newline}
    \newtext{#2}
}

\begin{document}

\title[Article Title]{\textit{Don't Push the Button!} Exploring Data Leakage Risks in Machine Learning and Transfer Learning}


\author[2]{\fnm{Andrea} \sur{Apicella}}\email{andapicella@unisa.it}
\equalcont{These authors contributed equally to this work.}
\author[1]{\fnm{Francesco} \sur{Isgrò}}\email{francesco.isgro@unina.it}
\equalcont{These authors contributed equally to this work.}
\author[1]{\fnm{Roberto} \sur{Prevete}}\email{roberto.prevete@unina.it}
\equalcont{These authors contributed equally to this work.}

\affil[1]{\orgdiv{Department of Electrical Engineering and Information Technology}, \orgname{University of Naples Federico II}, \orgaddress{\street{Via Claudio 21}, \city{Naples}, \postcode{80125}, \country{Italy}}}

\affil[2]{\orgdiv{Department of Information Engineering, Electrical Engineering, and Applied Mathematics (DIEM)}, \orgname{University of Salerno}, \orgaddress{\street{Via Giovanni Paolo II, 132}, \city{Fisciano (Salerno)}, \postcode{84084}, \country{Italy}}}




\abstract{\marklessfootnote{Published on Artificial Intelligence Review journal in Open Access available at \url{https://link.springer.com/article/10.1007/s10462-025-11326-3}. Please refer to the final published version having several improvements and typos corrections.}Machine Learning (ML) has revolutionized various domains, offering predictive capabilities in several areas. However, there is growing evidence in the literature that ML approaches are not always used appropriately, leading to incorrect and sometimes overly optimistic results. One reason for this inappropriate use of ML may be the increasing availability of machine learning tools, leading to what we call the ``push the button'' approach. While this approach provides convenience, it raises concerns about the reliability of outcomes, leading to challenges such as incorrect performance evaluation.
In particular, this paper addresses a critical issue in ML, known as data leakage, where unintended information contaminates the training data, impacting model performance evaluation. Indeed, crucial steps in ML pipeline can be inadvertently overlooked, leading to optimistic performance estimates that may not hold in real-world scenarios. The discrepancy between evaluated and actual performance on new data is a significant concern.
In particular, this paper categorizes data leakage in ML, discussing how certain conditions can propagate through the ML approach workflow. Furthermore, it explores the connection between data leakage and the specific task being addressed, investigates its occurrence in Transfer Learning framework, and compares standard inductive ML with transductive ML paradigms.
The conclusion summarizes key findings, emphasizing the importance of addressing data leakage for robust and reliable ML applications considering tasks and generalization goals.}




\maketitle
\section{Introduction}
Machine Learning (ML) has emerged as a revolutionary technology with vast potential across various domains \citep{jordan2015machine}. Its ability to extract patterns from data and make predictions has led to significant advancements in fields like healthcare \citep{shailaja2018machine,apicella2021high}, language translation \citep{lopez2008statistical}, and image interpretation \citep{ahmad2018interpretable,apicella2019contrastive}. 
\newtexttwo{As ML tools become increasingly accessible, in several domains it is often used as a ``black box'', focusing on achieving a desired outcome without delving into the details of how ML models act or the impact of specific choices on model performance. Multiple causes can lead to this type of approach. Among these, one can highlight the "big" availability of ML tools.} However, while the adoption of ML tools provides convenience and accessibility to ML applications, it may cause to overlook crucial considerations which can affect the reliability the outcomes generated. This approach can inadvertently lead to several issues, such as incorrect performance evaluation during the experimentation phase of ML systems and insufficient reproducibility  \citep{kapoor2023leakage,yang2022data}. Among these, a key issue is \textit{data leakage} - also known as \textit{pattern leakage} \citep{bouke2023empirical} - a problem where \textit{forbidden} information is unintentionally introduced into the training data, thus affecting model performance evaluation. More generally, data leakage can result in overly optimistic performance estimates during model development, which often fail to generalize to real-world scenarios. This typically occurs when, due to a lack of deep understanding, users unintentionally introduce forbidden information at various stages of the machine learning pipeline, such as during data preprocessing or feature selection. To illustrate the severity of the issue, \citep{rosenblatt2024data} point out that a high-profile study predicting suicidal ideation in youth showed no predictive power once feature selection leakage was corrected. Despite its methodological flaw, the original paper (now retracted) received 254 citations since its publication in 2017. 
The authors of \citep{wen2020convolutional} reviewed the validation procedures of $32$ papers of Alzheimer's Disease (AD) automatic classification with Convolutional Neural Networks (CNNs) from brain imaging data, highlighting that more than half of the surveyed papers may be affected by data leakage, and so the reported performance could be biased. Furthermore, the authors highlight that several papers lack of experimental details to infer if data leakage is present or not in the proposed studies. These examples underscores the critical importance of preventing feature leakage, which can sometimes occur in subtle and non-obvious ways.
\newtexttwo{However, while recent studies such as \citep{wen2020convolutional,yang2022data} have analyzed data leakage in ML , these analyses typically assume the classical inductive learning setting. However, this assumption does not always hold. In tasks framed as \textit{transductive learning} \citep{vapnik1999nature, joachims1999transductive}, for example, the model is designed to make predictions on a fixed, known set of instances. In such settings, what may appear as leakage under inductive assumptions can instead be valid, as will be discussed in detail in this work. }
\newtexttwo{Similarly, in \textit{Transfer Learning} (TL) \citep{yang2020transfer,apicella2022machine}, where data from different tasks or domains is reused during training, the boundaries between training and test distributions can be more complex. This can potentially lead to forms of leakage that differ from those in classical ML where inductive learning paradigm is adopted.}
\newtexttwo{Despite these challenges, most existing data leakage categorizations in literature do not take into account the learning paradigm (inductive vs. transductive), the ML framework (standard ML vs. TL), or the nature of the task. As a result, some types of leakage may go unnoticed or be misinterpreted.}

\newtexttwo{Therefore, differently from similar literature \citep{wen2020convolutional,yang2022data,kapoor2023leakage}, we aim at emphasizing the connections between data leakage, the specific task being addressed, the ML paradigm and framework adopted and data propagation through the learning and evaluation workflow. On the basis of this analysis, we propose a new and more comprehensive categorization for the data leakage problem, discussing how certain conditions causing data leakage can propagate through the ML pipeline. 
Summarizing, we offer the following main contributions: i) A new categorization of data leakage types in ML, focusing on the specific stages of the pipeline where leakage can occur; ii) we highlight the connection between the occurrence of data leakage and the specific task being addressed; iii) we investigate whether data leakage exists in TL frameworks, highlighting cases where it may or may not occur; iv) we explore how the occurrence of data leakage can be influenced by the type of ML learning paradigm used, specifically comparing inductive ML and transductive ML.}

\newtexttwo{The paper is structured as follows:  a summary and discussion of related works and existing data leakage categorizations proposed in the literature is provided in Sec \ref{sec:categorization}. In Sec. ~\ref{sec:background}, we provide background on core ML concepts, including differences between inductive ML, transductive ML, and TL, as well as an overview of a typical ML pipeline. Sec. \ref{sec:dataleaktypes} introduces a novel categorization of data leakage types, exploring, discussing in classical, transductive, and transfer learning contexts. Furthermore, we provide an empirical evaluation using synthetic data to illustrate the consequences of different types of data leakage. Finally, Sec. \ref{sec:final_disc} and \ref{sec:conclusion} offer concluding discussion and remarks about the key findings and insights presented in the paper.} 
For ease of reference, Table \ref{tab:acro} presents the most frequently used acronyms in this study.

\begin{table}[htbp]

\centering
\begin{tabular}{ll}
\toprule
\textbf{Acronym} & \textbf{Meaning} \\
\midrule
AD & Alzheimer's Disease \\
ANN & Artificial Neural Network\\
CNN & Convolutional Neural Network \\
DA & Domain Adaptation \\
DG & Domain Generalization \\
DL & Deep Learning\\
DNN & Deep Neural Network\\
EEG & Electroencephalography \\
ML & Machine Learning \\
TL & Transfer Learning \\
TML & Transductive Machine Learning \\
SDA & Supervised Domain Adaptation \\
XAI & eXplainable Artificial Intelligence \\
\bottomrule
\end{tabular}
\caption{Acronyms used in this work.}
\label{tab:acro}
\end{table}
\section{Related works}
\label{sec:related_works_data_leakage}

Data leakage is a common issue in classical ML that occurs when information out of the training data is unintentionally used during the model training. This phenomenon can lead to misleading results and evaluations, making the model appear more accurate than it actually is \footnote{see, for example, \url{https://imbalanced-learn.org/stable/common_pitfalls.html}}. Data leakage can be viewed as an instance of the \textit{circular analysis problem}, that happens when an analysis is based on data selected for showing the effect of interest or a related effect \citep{kriegeskorte2010everything}. A formal definition of data leakage is proposed in \citep{kaufman2012leakage}.

A considerable portion of research focusing on data leakage in machine learning is centered around specific domains or particular fields of study.
\newtexttwo{ For instance, \citep{bernett2024guiding} proposed a set of questions that should be asked to prevent data leakage in biological domains. Joeres et al. \citep{joeres2025data} proposed a software tool that aims to mitigate data leakage by performing data splits based on similarity or distance measures between data points, when such information is available. Whalen et al. \cite{whalen2022navigating} identify data leakage as a common pitfall in the application of machine learning to genomics.
The studies proposed in \citep{park2012flaws,li2020perils} highlight the potential pitfalls associated with designing suboptimal experimental setups in classification problems with Protein-Protein Interaction and electroencephalographic (EEG) data respectively. The authors emphasized how an imperfect experimental design could lead to wrong conclusions, subsequently influencing subsequent works in the field and biasing the established state-of-the-art, distorting the perceived advancements within the field. \newtexttwo{The study proposed by \citep{wu2021resting} highlights that many ML-based EEG studies suffer from small sample sizes and limited electrode coverage, leading to unreliable cross-validation results. Indeed, cross-validation is commonly performed on EEG samples rather than on participants. This practice leads to a situation where different samples from the same individual are included in both training and test sets, which contributes to inflated performance estimates. A similar point is also discussed in \citep{brookshire2024data}.} Similarly, the authors of \citep{kamrud2021effects} highlighted that a significant portion of EEG studies, which aim to classify the EEG of individual participants, often neglect to evaluate the trained models using unseen participants. This practice introduces a potential limitation in the generalizability and reliability of the ML for EEG classification since there is a risk that the models may be overfitted to the specific characteristics of the training group. The study by \citep{rosenblatt2024data} evaluates the impact of data leakage on connectome-based predictive models across four large datasets and three phenotypes. Testing over 400 pipelines, the authors investigate how leakage affects both prediction performance and model interpretation.}

More in general, \citep{kapoor2023leakage} surveyed prior literature reviews among $17$ scientific fields reporting that more than $290$ papers are affected by data leakage. It is interesting to notice that, among the $17$ fields inspected, $11$ are fields not directly related to computer science. Furthermore, the authors showed that, in several cases, repeating the reported experiments of several and well-affirmed studies with data leakage bugs corrected lead toward lower performance respect to the claimed ones, sometimes comparable with results given by simpler and older models.
\newtexttwo{a survey of the literature conducted in \citep{brookshire2024data} reveals that the majority of translational DNN–EEG studies adopt segment-based holdout strategies. As a result, most published DNN–EEG studies may substantially overestimate their classification performance on unseen subjects. Similar considerations were reported in \citep{varga2024exposing} about publised Wi-Fi signal's Channel State Information (CSI) Deep Learning systems.}
Instead, adopting software engineering methods, \citep{yang2022data} detected data leakage in about $30 \%$ of $100,000$ source code publicly available. However, the proposed analysis is made relying only on the source codes without considering the specific task addressed by each analyzed study. This can lead to consider several studies as affected by data leakage just because the publicly available source code contains pattern usually related with bad software practice, without considering ML paradigms where these patterns are legitimate. 
\newtexttwo{Dong et al. \citep{dong2022leakage} Investigated data leakage in models based on data from sports wearable sensors, proposing a Bayesian inference approach to detect leakage. The authors of  \citep{bouke2024empirical} conducted an empirical comparison of thirteen machine learning models for intrusion detection in 5G networks, with a strong emphasis on preventing data leakage during preprocessing.}

\paragraph*{Data leakage categorization in literature}
\label{sec:categorization}
To the best of our knowledge, there are few articles \citep{wen2020convolutional,yang2022data,kapoor2023leakage} proposing a taxonomy about data leakage types. The following of this section is dedicated to summarize them.
The authors of \citep{wen2020convolutional} identified four possible causes of data leakage in AD classification using CNNs: 1) wrong data split, happening when data from the same data source (e.g., the same patient) is used in more than a set between the training, validation, and test set; 2) late split: when data augmentation, feature selection or autoencoder (AE) pre-training are made before data split, conditioning the resulting validation/test sets; 3) Biased transfer learning: this category is specific for works involving transfer learning, and encloses the leakage  occurring when the source and target domains overlap. For example, when some subjects used to pre-train a model in the source domain are the same to evaluate the model in the target domain. 4) Absence of an independent test set: the test set is used also to evaluate the training hyper-parameters of the model and not only to assess the final model performance. Indeed, the paper highlighted that many of the surveyed studies did not motivate the adopted training hyper-parameter values or the chosen architectures, suspecting that the final models and reported performance were constructed relying on the test set.

\citep{yang2022data} provided an interesting data leakage categorization, distinguishing between at least four types of leakage: preprocessing, overlap, multi-test, and label leakages. Preprocessing leakage arises when training data and test data are preprocessed together, influencing each others. Overlap leakage can occur when test data are directly or indirectly (for example, during data augmentation or oversampling) used during the training. Multi-test leakage happens if data are used several times during model evaluation or hyper-parameter tuning, biasing the results. Finally, label leakage concerns when one or more input features should not be present in the training data since they directly map with the correct labels, conditioning the learning process. 

Differently, \citep{kapoor2023leakage} proposed a leakage taxonomy based on three macro-categories: 1) the lack of a clean separation between training and test dataset, 2) the adopted model uses features that are not legitimate, and 3) the test set used to evaluate the model is not drawn from the distribution of scientific interest. The first category consider the case when the training set and test set are not correctly divided, making the model able to access to information in the test data before the model performance is evaluated. The authors then consider as subcategories the cases 1.1) when training and test set are the same, 1.2) when preprocessing on the whole dataset can result in a leakage of information from the test to the training set, 1.3) when feature selection is made on the whole dataset, providing information on which features perform well on test set, and 1.4) when there are duplicate data between training and test set. The second macro-category consider scenarios when a ML model uses features which should not be adopted for the task at hand, since they can be proxies for the correct output. The third macro-category considers the case when the evaluation is made on data outside from the distribution of specific interest. The subcategories considered are 3.1) the temporal leakage, 3.2) Non-independence between training and test samples, and 3.3) sampling bias in test distribution. Temporal leakage 3.1) happens when the test set contains data temporally earlier to the training data. In this case, training data could contain information taken from the future respect to the test one, which can help the classification of earlier data. 

Possible actual instances of data leakage which can be classified as temporal leakage  (subcategory 3.1 of \citep{kapoor2023leakage}) are discussed in \citep{plotz2021applying} for sensor data analysis and in \citep{lyu2021empirical} for AIOps with time related data. In particular, sensor data discussed in \citep{plotz2021applying}, being time series data, are usually segmented into overlapping frames and analyzed individually. However, in this case, consecutive frames are not independent of each other, violating the identically and independently distributed (i.i.d.) condition needed in several validation frameworks (such as cross validation). Furthermore, in cases of overlapping frames, identical parts of the signal can be distributed into more than one frame. The authors of \citep{lyu2021empirical} argued that employing a random split between training and out-of-training data in AIOps with time-related data led to a misleadingly higher model performance compared to using a time-related data split, where only historical data is data during the training phase. However, the proposed study consider only AIOps, without generalizing to other ML domains. 
Regards the last two subcategory proposed in \citep{kapoor2023leakage}, the subcategory 3.2  (Non-independence between training and test samples) can happen for example in medical studies when the same patient provides data to both the training and the test data. However, this case is directly related to the constraints of the scientific claim. Finally, the subcategory 3.3 regards sampling bias in test distribution, which happens when data sampled for test data are not representative of the population of interest. 

\paragraph*{Current shortcomings and our contributions}
The categorizations proposed in \citep{wen2020convolutional,yang2022data,kapoor2023leakage} seems to mix together the concept of information leakage and evaluation bias. For example the multi-test leakage defined in \citep{yang2022data}, corresponding to the absence of an independent test set in \citep{wen2020convolutional}, is an instance of the well known Multiple Comparisons Problem \citep{neter1996applied}. As such, it should be considered as a form of bias in the ML model evaluation more than a proper form of information leak. Indeed, there is not an actual leakage of knowledge from the evaluation/test data to the training stage, but a bias in the evaluation step due to multiple comparison on the same data. Similar considerations can be made for the leakage subcategory 3.3 (that is "sampling bias in test distribution") proposed in  \citep{kapoor2023leakage}. Also in this case, this category regards problems related to the test data distribution which can lead to biased results in the model evaluation, which is a different concept respect to the data leakage.
In other words, data leakage can lead to biased evaluations, but a biased evaluation is not necessarily caused by a data leak. Instead, bias in the data can lead to data leakage. 

Notably, the greatest part of these works does not take into account the task addressed. Among the inspected works, the only one which hints to the task is \citep{kapoor2023leakage}, which invited the reader to consider the scientific claim to understand if there is data leak or not. Furthermore, all these categorizations refer to classical supervised ML. Instead, different ML frameworks, such as transductive learning and transfer learning are not addressed. Therefore, Data Leakage depends not only by the task at hand and the overlap between the data, but also on the adopted ML framework. The only categorizations which takes into account a different ML framework was \citep{wen2020convolutional} where the biased transfer learning case is considered.

\section{Background}
\label{sec:background}
In this section, to better understand the risks of data leakage, we will provide some basic ML concepts and which is the pipeline, although different ones can be used depending on the case, for the evaluation and construction of the final ML model in a standard ML approach. In addition, we will discuss some important ML paradigms such as Transductive ML (TML) and Transfer Learning (TL) which have key differences with respect to the basic assumptions of standard ML approaches and thus determine different types of data leakage problems.

ML encompasses various methodologies, broadly categorized into several types based on the nature of the learning process, which include supervised, semi-supervised, and unsupervised ML, and reinforcement learning. In this work, we will focus only on the supervised and semi-supervised approaches:  supervised ML the available data is paired with corresponding output labels, while Semi-supervised ML exploits only labeled data, while semi-supervised ML leverages both labeled and unlabeled data to improve model performance.

\subsection{A classical Machine Learning pipeline}
\label{sec:background:pipeline}
To better understand the nuances and risks associated with data leakage described and discussed in the subsequent sections, in this section the main steps of a classical ML pipeline are briefly described. Notice that a standard pipeline for ML processes is not commonly shared in the literature, although there are a number of proposals in this regard \citep{shearer2000crisp,schlegel2023mamagement,chaoji2016machine,amershi2019software,salvaris2018deep}. We build upon the key concepts and suggestions from \citep{ashmore2021assuring} to outline a standard pipeline.

\begin{figure}
    \centering
    \scalebox{0.8}{\tikzset{every picture/.style={line width=0.75pt}} 
    \begin{tikzpicture}[x=0.75pt,y=0.75pt,yscale=-1,xscale=1]
\input{_TKZ_PIPELINE}
\end{tikzpicture}}
    \caption{A classical ML supervised pipeline.
    See text for further details.
    }
    \label{fig:supervised_pipeline}
\end{figure}

Following \citep{ashmore2021assuring}, the production of a ML model usually involves three main stages: \textit{data management}, \textit{model learning}, and model verification (which we will refer to as \textit{model evaluation} for better clarity). Data management includes the acquisition of data from the given sources and all the processing steps that lead to the final dataset for the learning and evaluation stage.
In the model learning stage, the model learns patterns and relationships from the provided final dataset. Finally, during the evaluation stage, the trained model is evaluated considering one or more performance measurements. More deeply, during the training stage, the model tunes its inner parameters to minimize a defined loss or error function. 
The general goal is for the model to learn patterns from the training data to make accurate predictions on data not used during the training, i.e. out-of-the-training data. In standard ML setting, once the model has been trained, the performance of the final model are assessed on a dataset composed of data not seen by the model during the training (usually referred as \textit{test set}) to evaluate how well the trained model performs on new, unseen instances. It is important to highlight that  the ``test set'' if often named as ``validation set'' in literature. However, in this study we indicate the set used to evaluate the model after the training as ``test set'', reserving the name ``validation set'' for other purposes (see Sec. \ref{sec:preprop:particular}).  More in general, as stated in Sec. \ref{sec:background:notation}, we will refer to every dataset used to evaluate the model as \textit{evaluation set}. 

Each stage encompasses multiple steps, as illustrated in Fig. \ref{fig:supervised_pipeline}. The following of this section provides a discussion and description of these individual steps on key aspects which are involved in our study. For further details see \citep{ashmore2021assuring}.

\paragraph{Data Management} 
The authors of \citep{ashmore2021assuring} include in this stage four key activities, that are collection, augmentation, preprocessing, and analysis. 

\begin{figure}
    \centering
    \scalebox{0.8}{\tikzset{every picture/.style={line width=0.75pt}} 

\begin{tikzpicture}[x=0.75pt,y=0.75pt,yscale=-1,xscale=1]
\input{_TKZ_PIPELINE_DATA_MANAGEMENT}

\end{tikzpicture}}
\caption{In violet he steps involved in the data management stage in a classical ML supervised pipeline. See text for further details.}
\label{fig:supervised_pipeline_data_management}
\end{figure}

However, we want to emphasize that the process of data collection should also take into account both i) how the data are distributed, ii) how the the data are designed in terms of composition (which we will refer as \textit{data design}), and iii) the specific ML task being considered. In fact, as we will discuss below, these factors can significantly impact on the model evaluation, potentially leading to data leakage conditions. \newtexttwo{In Fig. \ref{fig:supervised_pipeline_data_management} the main pipeline steps involved during the daata management stage.}
Augmentation and pre-processing play different roles in the enhancement of the dataset: augmentation supplements the data 
by generating additional instances or reducing the existing ones, while preprocessing works on existing data instances. More in general, in this paper we will consider as part of the Data Management stage all the operations involved in generating, reducing or updating the data instances, such as oversampling, undersampling, and imputation, which are aimed at constructing a final dataset that is ready to be used for the learning and evaluation stage of the ML model. In particular, we will refer to \textit{instances synthesis} as all the operations related to adding or removing of new instances after the data collection phase, while we will refer to \textit{preprocessing} as all the operations and transformations applied to data before it is used in a ML model, 
such as reducing noise, handling missing values (imputation), and standardizing features. 
The final output of the Data Management stage is the data involved in the ML workflow, that are a training set $D^T$ and out-of-the-training  sets. Usually, one or more out-of-the-training sets are used to evaluate the model (referred to as verification set in \citep{ashmore2021assuring} or evaluation set). Among the evaluation sets, the test set ${D}^E$ has to be completely unseen during training, so that it can provide an unbiased evaluation of the model behavior on new data.
We will refer to the procedure used to decide how to split the dataset into training and evaluation sets as the \textit{split strategy}. 

\paragraph{Model Learning} In \citep{ashmore2021assuring} the Model Learning stage consists in  creating a ML model from the data presented to it. 
\begin{figure}
    \centering
    \scalebox{0.8}{\tikzset{every picture/.style={line width=0.75pt}} 
\begin{tikzpicture}[x=0.75pt,y=0.75pt,yscale=-1,xscale=1]
\input{_TKZ_PIPELINE_LEARNING}
\end{tikzpicture}}
\caption{In fuchsia the steps involved in the learning stage in a classical ML supervised pipeline. See text for further details.}
\label{fig:supervised_pipeline_learning}
\end{figure}

However, we want to highlight that, before the effective model training, a feature engineering step is usually made in several ML pipelines. With \textit{feature engineering} in ML we mean the process of selecting, transforming, or constructing features from existing data that help the ML algorithm to find the underlying patterns or relationships within the data. \newtexttwo{In Fig. \ref{fig:supervised_pipeline_learning} the main pipeline steps involved during the model learning stage.}
Usually, feature engineering involves two main sub-steps: \textit{feature extraction}, consisting in  generating new features by transforming or combining existing features, and \textit{feature selection}, consisting in identifying and choosing the most relevant and informative features from the original data. 
Usually, feature engineering may involve performing either feature selection, feature extraction, or a combination of both. 
Feature engineering and preprocessing are not strictly sequential processes and can occur interchangeably or in various orders based on the characteristics of the data and the task. Furthermore, it is important to highlight that Deep Neural Networks (DNNs) can automatically learn hierarchical representations from data, reducing the necessity for explicit feature engineering steps. However, there might still be scenarios where preprocessing could be beneficial, even within deep learning frameworks.
Furthermore, several operations similar to preprocessing steps like normalization or scaling are often integrated into the network architecture or conducted as part of the training pipeline itself (see for example \cite{ioffe2015batch}).  
After any feature engineering methods have been applied to ${D}^T$, the resulting data can be fed to the training stage, where the model parameters are learned. 

\paragraph{Model evaluation} The evaluation stage involves assessing the performance of a trained model using evaluation sets (such as the test set ${D}^E$) not encountered during the training process. 
\begin{figure}
    \centering
    \scalebox{0.8}{\tikzset{every picture/.style={line width=0.75pt}} 

\begin{tikzpicture}[x=0.75pt,y=0.75pt,yscale=-1,xscale=1]
\input{_TKZ_PIPELINE_EVALUATION}
\end{tikzpicture}}
\caption{In orange the steps involved in the evaluation stage in a classical ML supervised pipeline. See text for further details.}
\label{fig:supervised_pipeline_evaluation}
\end{figure}

\newtexttwo{In Fig. \ref{fig:supervised_pipeline_evaluation} the main pipeline steps involved during the model evalution are shown.}
It's important to emphasize that the preprocessing and feature engineering steps applied to the $D^T$ must also be consistently applied to the evaluation tests. This consistency ensures that the evaluation data follows the same preprocessing and feature engineering procedures as the training data, preserving the integrity of the evaluation process. 
Finally, the model predictions on the evaluation set are compared against the ground truth values (for example, the actual labels in classification problems), and one or several performance metrics are computed to evaluate the trained model. 

\subsection{\newtext{Generalization in Machine Learning: Inductive, Transductive, and Transfer Learning}}
\label{sec:background:indvstrans}

During the ML evaluation stage, what has to be effectively evaluated depends on the ML paradigm adopted. We can distinguish between two typical paradigms: \textit{inductive} ML setting and \textit{transductive} ML setting \citep{vapnik1999nature}. Inductive ML is the more prevalent setting and involves training a model on a dataset to generalize and make predictions on new, unseen data. In this case, the focus is on the model's ability \textit{to generalize} to previously unseen data instances and validation procedures, such as Cross Validation or Hold-Out, should consider these requirements. In other words, we need that the model is able to generalize on unseen, but also \textit{plausible and task-related} data.
Conversely, transductive ML \citep{vapnik1999nature} aims for high performance \textit{specifically on the available, pre-existing data}, without generalization purposes. In other words, generalization to new, unseen data instances is not the goal, and the model's effectiveness is localized to the current available data. 
\newtexttwo{Summarizing: 
\begin{itemize}
    \item \textbf{Inductive Learning} focuses on generalization to new, unseen data. This is the standard ML setting.
    \item \textbf{Transductive Learning} aims to make predictions only on the specific test data available during training.
\end{itemize}} 

In many cases of transductive or inductive paradigms, we assume that we are dealing with problems involving one dataset domain \cite{yang2020transfer}. By contrast, Transfer Learning (TL, \cite{yang2020transfer}) approaches assume that data from different domains are available during the training phase of a model. \newtext{Indeed, TL leverages knowledge from different domains to enhance model performance. Several TL strategies have been proposed and categorized into two main subfamilies within the literature, that are Domain Adaptation (DA) strategies and Domain Generalization (DG) strategies. DA strategies focus on adapting a model from one or several source domains to a target domain by minimizing domain shifts or differences between them. DA usually assumes that an unlabeled dataset ${D}_{target}$ from a target domain is available during the training stage, together with labeled data ${D}_{source}$ from one (or more) source domain(s).} A subset of DA strategies, known as Supervised DA (SDA), utilizes only labeled data from both the source and target domains during training. For example, fine-tuning methods exploit models trained on datasets in a source domain and fine-tuning them using data from a specific task on a particular target domain.
\newtext{Conversely, DG strategies aim to extract knowledge from multiple source domains without access to the target domain, emphasizing the creation of models that generalize well to new unseen domains.} DG assumes that several dataset $\{{D}_{source_j}\}_{j=1}^S$ acquired from different source domains are available, and no data on the target domain is known.
\newtexttwo{Summarizing: 
\begin{itemize}
    \item \textbf{Domain Adaptation (DA)}: access to labeled source data and unlabeled (or partially labeled) target data.
    \item \textbf{Domain Generalization (DG)}: no access to target data during training; aims for generalization across domains.
\end{itemize}}
Evaluation data for TL-based systems must closely reflect the target domain and the task addressed, requiring careful selection.
Furthermore, as in classical ML, a key aspect to be considered in TL scenarios is whether the examined problem fits into an inductive or transductive\footnote{It is noteworthy, however, that the term "transductive" has frequently been misapplied in TL literature (see \citep{moreo2021lost}). 
In line with \citep{moreo2021lost}, in this work the term "transductive"  is adopted as in the original Vapnik formulation \citep{vapnik1999nature}, even in the context of TL scenarios.} ML setting. In simpler terms, one must ascertain whether the task's goal is to generalize on new data or merely to accurately predict the available data in the target domain(s). Also in TL, we distinguish between ``transductive'' and ``inductive'' cases, the former adopted if the aim of the TL framework is to make correct prediction only on the available data in the target domain(s), the latter if the aim is to generalize toward target domain(s) with unavailable data during the training stage. Similarly to the ML case, determining if the problem requires a transductive or inductive setting can determine if a leakage condition is present or not.

\section{Data leakage: definition and categorization}
\label{sec:dataleaktypes}

\subsection{Used notation and basic definitions}
\label{sec:background:notation}
In this work we will use the following notation: in the case of supervised ML, a dataset is a set of pairs ${D}=\{(\vec{x}^{(i)}, \vec{y}^{(i)})\}_{i=1}^n \in \mathbb{D}$ where each $\vec{x}^{(i)}$ is a data instance which is processed by the ML model to return a prediction, $\vec{y}^{(i)}$ represents the ground truth or the actual label value that the model aims to predict or estimate when provided with the input $\vec{x}^{(i)}$, and $\mathbb{D}$ is the set of all the possible datasets for the given task. Together with the term ``target'', the term \textit{label} is often used in machine learning to refer to the desired output of a system. In this work, to avoid confusion with the term ``target'', which is frequently used in different context such as TL (see Sec.\ref{sec:background:indvstrans}), we adopt ``label'' specifically to denote the desired output of a machine learning system, while reserving the term "target" for use in other cases. We mean with \textit{model} a trainable ML model $M \in \mathbb{M}$ where $\mathbb{M}$ is a family of possible ML models (e.g., Artificial Neural Networks, SVMs, etc.).
Usually, $\vec{x}^{(i)}\in \mathbb{R}^d$ where $d$ is the dimensionality of the data instance, and $\vec{y}^{(i)}\in \mathbb{R}^c$ where $c$ is the dimensionality of the output. For the sake of simplicity and without loss of generality, in this work we consider the labels as one-dimensional values, that is $y^{(i)}\in \mathbb{R}$. 

\newtext{We will distinguish between a training dataset $D^T$, which contains the data used to train a model $M$, and an out-of-the-training dataset, such as an evaluation dataset $D^E$, which is used to assess the performance of the trained model. \newtexttwo{Finally, similarly to \citep{izmailov2022feature}, with the term \textit{spurious} we refer to any information, such as a feature in $\vec{x}^{(i)} \in \mathbb{R}^d$, containing patterns correlated with the task label of the data but not inherently relevant to the task (for example, the image backgrounds when classifying the foregrounds). We emphasize that the property of a feature to be spurious or not strongly depends on the task. For instance, the background of an image may be considered spurious in a classification task focused on identifying foreground objects. Conversely, if the goal is to distinguish between images taken during the day or at night, background features become relevant and informative. In this work,  we define the predicate $Spurious_T(x^{(i)}_j, y^{(i)})$ to be true when the feature $x^{(i)}_j$ is spurious for the task $T$ carrying unintended or shortcut information about the label $y^{(i)}$ for the task $T$, false conversely. In a similar way, we can define  the predicate $Spurious_T(x^{(i)}_j, x^{(u)}_h)$ to be true if the $x^{(i)}_j$ carries  information about the feature $x^{(u)}_h$ of another instance $j\neq u$, in a way that does not reflect a real-world relation that the task is designed to learn or exploit. For instance, in a classification task where each instance represents an independent patient, the blood pressure of patient $i$ should not contain information about the heart rate of patient $u$. If such information is introduced in some way, the model may exploit unintended statistical dependencies between different patients, which do not reflect the real-world structure of the task.
Finally, we can say that a dataset $D$ is spurious for the tast $T$, i.e. $Spurious_T(D)$ is true, if it exists at least a data in $D$ containing spurious features (respect to the label or other data instances) for the task $T$. Finally, we distinguish between a feature that \textit{naturally assumes} a value from an observation or a measurement and a feature that is \textit{assigned} a value. Specifically, we use the notation $x_j^{(i)} = v$ to indicate that the $j$-th feature of the $i$-th instance has the value $v$ as observed during the data acquisition. Conversely, we use $x_j^{(i)} \leftarrow v$ to denote that the value $v$ is assigned to the feature $x_j^{(i)}$, for example, by design or construction or during a preprocessing operation.}
\newtexttwo{Importantly, whether a dataset $D$ is affected by such leakage depends on the task $T$:
the same dataset might be spurious with respect to one task and perfectly valid for another.
Hence, $spurious_{T_1}(D)$ is true does not imply $spurious_{T_2}(D)$ is also true for $T_1 \neq T_2$.
Finally, we distinguish between a feature that \textit{naturally assumes} a value from an observation or a measurement and a feature that is \textit{assigned} a value. Specifically, we use the notation $x_j^{(i)} = v$ to indicate that the $j$-th feature of the $i$-th instance has the value $v$ as observed during the data acquisition. Conversely, we use $x_j^{(i)} \leftarrow v$ to denote that the value $v$ is assigned to the feature $x_j^{(i)}$, for example, by design or construction or during a preprocessing operation.
}
}

\subsection{Proposed taxonomy}
\begin{figure}
    \centering
\scalebox{0.85}{\begin{tikzpicture}
  \node at (0, 12) {
    \parbox{1.0\textwidth}{
      \scalebox{0.8}{\input{TAB_SUMMARY_TREE1}}
    }
  };
  \node at (0, 8.5) {
    \parbox{1.0\textwidth}{
      \noindent\makebox[\textwidth]{\textcolor{blue}{{\scriptsize \textbf{During the dataset definition}}}}
      \textcolor{blue}{\hdashrule[0.5ex]{\textwidth}{1pt}{3mm 2mm}}
    }
  };
  \node at (0, 4) {
    \parbox{1.0\textwidth}{
      \scalebox{0.8}{\input{TAB_SUMMARY_TREE2}}
    }
  };
  \node at (0, 0.5) {
    \parbox{1.0\textwidth}{
      \noindent\makebox[\textwidth]{\textcolor{blue}{{\scriptsize \textbf{Before data split}}}}
      \textcolor{blue}{\hdashrule[0.5ex]{\textwidth}{1pt}{3mm 2mm}}
    }
  };
  \node at (0, -2) {
    \parbox{1.0\textwidth}{
      \scalebox{0.8}{\input{TAB_SUMMARY_TREE3}}
    }
  };
  \node at (0, -4.5) {
    \parbox{1.0\textwidth}{
      \noindent\makebox[\textwidth]{\textcolor{blue}{{\scriptsize \textbf{Before model training}}}}
      \textcolor{blue}{\hdashrule[0.5ex]{\textwidth}{1pt}{3mm 2mm}}
    }
  };
\end{tikzpicture}}
    \caption{\newtext{Data leakage categorization with respect to the considered ML pipeline. Each edge represents a possible type of leakage, highlighting reference where examples of this type of leakage are made and/or discussed. The blue dashed lines indicate the temporal points in the ML pipeline where the leakage risks are introduced, emphasizing the importance of addressing these concerns at different stages of the ML process. Citations on the edges refer to studies that are either affected by or discuss the corresponding type of leakage.}}
    \label{fig:leakage_type}
\end{figure}
\newtext{
First of all, it should be emphasized that a formal definition of data leakage is missing in the literature. Although there are some attempts in this sense (e.g. \cite{kaufman2012leakage}), they are not widely shared. More informally, we can say that data leakage occurs when a ML model is trained using information that should not be available during training, thereby altering the evaluation process. Based on these considerations, we will give a new perspective on data leakage, defining it on the basis of the ML pipeline introduced above, with the idea that it can occur at different stages of this pipeline.
As we will show in the following, understanding whether the task at hand requires generalization is fundamental when determining whether a model $M$ is affected by data leakage and at which step of the pipeline is present. Indeed, data leakage becomes problematic specifically in contexts where the model is expected to generalize well to new, unseen data (such as in Inductive ML). If the goal is purely to perform well on available data (e.g. transductive ML) without regard for performance on new data, the issue of leakage may not be critical.
At a high level, referring to a classical ML pipeline as described in Sec. \ref{sec:background:pipeline}, we have identified three critical points:} 1) During the dataset definition phase, issues related to how data is collected or generated might inadvertently lead to data leakage.  
2) Before separating data into training and out-of-the-training sets, data preprocessing might unintentionally introduce leakage. 3) After the data is split, there's a risk of data leakage due to the way the data are split, potentially resulting in an inflated assessment of model performance.
\newtext{In relation to these points, we can identify three categories of leakage: 
\begin{enumerate}[i.]
    \item \textit{data-induced leakage}: when spurious information is embedded in the data already during the dataset construction;
    \item \textit{preprocessing-related leakage}: when a preprocessing procedure is used without considering separation between training and evaluation datasets;
    \item \textit{split-related leakage}: when information leaks between training and evaluation datasets during the split between them.  
\end{enumerate}
}
In Fig. \ref{fig:leakage_type} we report the types of leakage inspected in this work, together with a descriptive summary of each type of data leakage is provided, along with references to studies addressing each specific type.
In the next subsections we will discus in detail these three types of data-leakage. \newtext{For each subsection, we present possible scenarios where these types of leakage can arise in classical inductive ML settings, with several examples drawn from the existing literature. We then explore scenarios where such cases may occur in different ML paradigms, such as transductive ML and TL, providing analysis and discussion. Finally, a summary discussion for each subsection is provided.}



\subsection{\textbf{Data-induced leakage}}
\label{sec:dataleaktypes:inner}
During the process of defining a dataset, the components that constitute the input of the ML pipeline are carefully selected. However, it becomes crucial to ensure that the features composing the input of the ML pipeline do not contain artificial ``spurious'' information which can lead the model toward the correct label not by using characteristics and pattern of the task under consideration, but specific and spurious characteristics of the specific adopted dataset. 
In other words, the leakage originates during the process by which $D$ was selected, aggregated, or designed.
Indeed, spurious information could be exploited during the training process leading the classifier toward the correct output for each input, but using information which should not be available for the task in real scenarios.

\newtext{We identified the following Data-induced leakage cases: 
\begin{enumerate}[i.]
    \item \textit{Data collecting leakage}: when the process of acquiring the training data introduces spurious information or unintended correlations that influence the model's learning.
    \item \textit{Label leakage}: when the label labels are part, directly or indirectly, of the chosen data representation features. 
    \item \textit{Synthesis leakage}: When both the training and evaluation datasets are generated by a function $g: \mathbb{D} \to \mathbb{D}$ applied on the available data, influencing either the training phase, the evaluation phase, or both, by accessing to forbidden information.
\end{enumerate} In the following of this subsection, we will describe them more deeply.}

\subsubsection{Data collecting leakage} 
\label{sec:dataleaktypes:inner:collect}
We can consider this kind of leakage when information leakage appears during the data acquisition process, i.e. if it is caused by the acquisition process by which $D$ was obtained. 

\newtext{In the following of this section we will describe and discuss three possible scenarios of this kind of Data collecting leakage.}

\casestudybox{A Data collecting leakage due to the data acquisition sources}{Consider a dataset $D$ collected from $S$ subjects with differing levels of expertise, i.e. $D=\bigcup\limits_{s=1}^S D_s$ where each $D_s$ is the acquisition set from the subject $s \in \{1,2,\dots, S\}$. Depending on the task under investigation and the type of data collected, different cases are possible. For example, when the involved subjects possess a deeper understanding or expertise in the task, the resulting acquisitions might result in more accurate or reliable data. Conversely, the same condition can lead to addiction to the stimuli involved in the data generation, leading to data where the informative part for the task at hand is lost. Therefore, failing to account for the subjects' expertise could result in a model trained on data $D^T \subset D$ too similar between them due to subjects with similar expertise. This user expertise can be a form of undesirable knowledge shared between training data and evaluation data, biasing the model performanceon an evaluation set $D^E\subset D\setminus D^T$. For instance, when both the training and evaluation sets include data from subjects with similar expertise levels, the model could overfit to those specific patterns.}
However, in this case the existence of leakage depends on the task's goals: if the final model is meant to be used only by subjects with similar expertise levels, this similarity might be acceptable. Conversely, if the model is intended to generalize to users with a wide range of expertise, the failure to account for expertise differences could lead to data leakage, as the training and evaluation sets are not independent in terms of the characteristic (i.e. expertise) which can result relevant for the trained model. 

\casestudybox{A data collecting leakage due to the data acquisition design}{In \citep{spampinato2017deep} an EEG classification task involving visual object stimuli was proposed. The adopted data acquisition employed a methodology in which all stimuli belonging to a particular class were presented consecutively (``in block'') to the subjects. This experimental block design, together with the split strategy adopted, has the consequence that every trial in each evaluation set comes from a block that contains many trials in the corresponding training set. Therefore, the resulting reported high task performance could not be due to capturing stimulus-related brain activity, but identifying temporal artifacts within the data. Indeed, \citep{li2020perils} repeated the same experiments but with new data gathered using two different paradigms: the former adopting again a block design, 
and the latter adopting a rapid-event design, where the stimuli were presented in randomized order. The results showed that high performance resulted with the block design adopted cannot be replicated with a rapid-event design. 
}

\casestudybox{A data collecting leakage due to data aggregation}{
The review \citep{roberts2021common} highlighted the risks of datasets used in several studies to diagnose and prognosticate COVID-19 through chest computed tomography and chest radiography data. In particular, the adoption of datasets composed of data coming from other ones (\textit{Frankenstein datasets}) and redistributed under a new name seems to be a common practice across several studies in literature. This  could lead to the presence of repeated or overlapping data within the aggregated dataset, since one or more datasets composing the Frankenstein dataset can in turn contain other datasets which can be part of the Frankenstein dataset. \newtexttwo{However, Frankenstein datasets are typically not assembled as unions of subsets, and duplicates may be allowed. As a result, it is possible that two datasets $D_{s_1}$ and $D_{s_2}$ composing $D$ satisfy $D_{s_1} \cap D_{s_2} \neq \emptyset$. Consequently, models trained and evaluated on supposedly distinct datasets, such as $D^T\subseteq D$ and $D^E \subseteq D$ respectively, might inadvertently be using identical or overlapping data, wrongly assuming them to be different. } 
\newtexttwo{
In Fig. \ref{fig:frankenstein} an example of this kind of scenario is shown.} 
\begin{figure}
    \centering
    \includegraphics[width=0.8\linewidth]{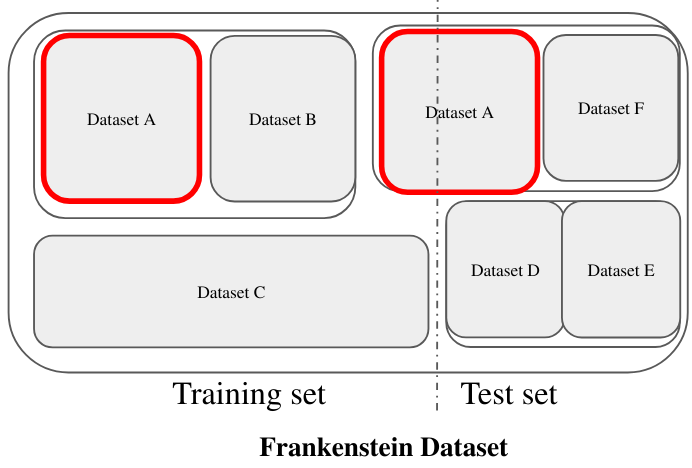}
    \caption{\newtexttwo{An example of a Frankenstein dataset composed of several datasets, where some data instances are shared between training and test set.}}
    \label{fig:frankenstein}
\end{figure}

}


\subsubsection{Label leakage} 
\label{sec:dataleaktypes:label}
As said at the beginning of this section, we consider a data leakage condition as \textit{label leakage} when the process of data definition makes the labels as part, directly or indirectly, of the chosen data representation features. This practice leads the ML model to over-fit on the training data, as it can access to the correct label during training. Consequently, during model evaluation the model might appear to perform exceptionally well by extracting the label from the features themselves.
 \newtext{More precisely, we can distinguish between:
\begin{enumerate}[i.]
\item \textit{Direct label leakage}: when data representations contain features corresponding to the correct label;
\item \textit{Indirect label leakage}: when data representations contain features that, while not explicitly encoding the label, are spurious, or capture effects or decisions influenced by \textit{a priori} knowledge of the real label. 
\end{enumerate}
}

\paragraph{Direct label leakage} 
In this case, actual labels are part of the data representation features used as input for the ML model, i.e. 
\begin{definition} \newtexttwo{Direct label leakage can happen if $\exists i\in\{1,\dots,|D|\}, f \in \{1, \dots,d\} \text{ s.t. }x^{(i)}_f \leftarrow y^{(i)}$.}
\end{definition}
In other words, a direct label leakage happens when $Spurious_T(D)$ is true due to data representations $\vec{x}^{(i)}$ in $D$ when the correct label is assigned for the task $T$ as value of a feature.
In this case, the ML model could exploit the label coinciding to the feature $x^{(i)}_f$, resulting in an artificially high evaluation of the model's performance.
Moreover, direct inclusion of the label within the features poses a critical challenge when dealing with new unlabeled data. Since the model depends on the label being readily available as a feature, it becomes impossible to  apply this learned model to new, unseen data lacking this explicit label within the features. 

\paragraph{Indirect label leakage} 
We consider as indirect label leakage when information regarding the correct labels are integrated into the features but in an indirect way, for example as one or more proxy features computed from the correct label. More formally
\begin{definition} \newtexttwo{we can have indirect label leakage when: $\exists i \in \{1 \dots |D| \}, f_1, f_2, \dots, f_t \in \{1, \dots,d\}, k : \mathbb{R} \longrightarrow \mathbb{R}^t$, s.t $\vec{x}^{(i)}_{f_1, f_2, \dots, f_t} \leftarrow k(y^{(i)})$}
\end{definition}



Notice in the definition the presence of a function $k$ affecting the value of a subset of features. This highlights that certain features in $\vec{x}^{(i)}$ are not independent from the label $y^{(i)}$, but are instead derived from a relationship $k(y^{(i)})$. These features, therefore, capture information that is influenced by the \textit{a priori} knowledge of the label.

The following of this section shows and discusses some scenarios of indirect label leakage.

\casestudybox{An indirect label leakage due to spurious feature}{In \citep{rosset2010medical} two data mining competitions in medical tasks are  Discussed. The adopted datasets consist in features acquired by several patients with a given disease. In particular, in one dataset the patient identifiers (IDs) were provided as a feature of the data instances, i.e. for each data point $\vec{x}^{(i)}, \exists f$ s.t. $x_f^{(i)}=ID$, i.e. $Spurious_T(\vec{x}^{(i)}_f,y^{(i)})$ is true, carrying information towards identifying patients with malignant candidates. In this context, its relevance to predicting the outcome might not directly derive from the medical condition itself, but could be due to, for example, regional or demographic factors. Therefore, the ID is a spurious feature carrying indirectly information about the correct label (e.g., the majority of data instances belonging to one class have a low ID number). 
This scenario is graphically illustred in Fig. \ref{fig:label_leak}
\begin{figure}
    \centering
\includegraphics[width=0.8\linewidth]{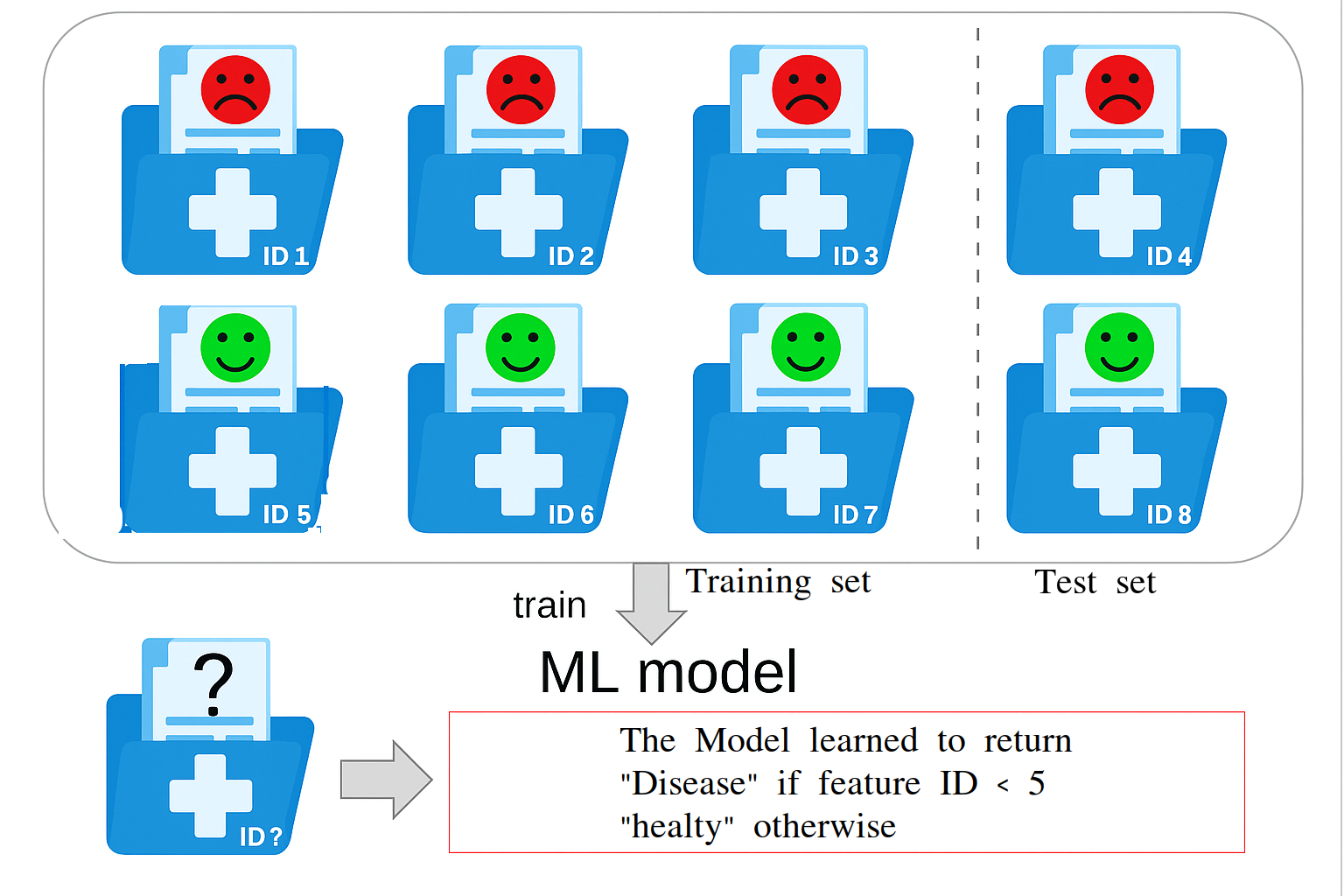}
    \caption{\newtexttwo{An Indirect Label leaking scenario. The ID is a spurious feature carrying indirectly information about the correct label.}}
    \label{fig:label_leak}
\end{figure}
}

\casestudybox{An indirect label leakage due to priori knowledge}{The authors of \citep{ye2018prediction} proposed a ML model for predicting hypertension risk based on predictors extracted from the electronic health records of several patients. 
While the proposed model achieved strong results, the authors of \cite{chiavegatto2021data}, upon reviewing the key variables listed in the appendix of \citep{ye2018prediction}, noted that several of the most influential predictors were, in fact, common antihypertensive drugs.
}
It is evident that information regarding the correct labels are subtly integrated into the features but in an indirect way. \newtext{In particular,the most influential features are based on decisions made with prior \textit{knowledge} of the correct class, creating a sort of circular dependency where these features cannot be meaningfully exploited without already knowing the class label.}

\subsubsection{Synthesis leakage} 
\label{sec:dataleaktypes:synthesis}
With Synthesis leakage we identify the cases when leakage is due to the generation of the actual data which will be processed by the ML pipeline. It often arises when there are issues with data balancing, such as over-represented or under-represented classes. More formally, let a procedure $g: \mathbb{D} \to \mathbb{D}$ whose generate the dataset $D_{gen}=g(D)$ and let $D^T \subseteq D_{gen}$ and $D^E\subseteq D_{gen}$. A synthesis leakage can occur if $\exists (\vec{x}^{(i)}, y^{(i)}) \in D^T, (\vec{x}^{(j)}, y^{(j)}) \in D^E$ are generated by $g$ from the same $D'\subseteq D$.


\newtext{In other words, there exist instances in $D^E$ and $D^T$ generated by using knowledge from the same subset of $D$ and resulting in task-spurious information in $D_{gen}$. The function $g$ applied to the original data can lead $D^E$  and $D^T$  to influence each other, leading to a situation where information from one set unintentionally leaks into the other. This situation occurs when the generation function is applied to the data before splitting it into training and evaluation sets. In fact, generation functions may be improperly used during the data construction process, prior to the selection of the training and evaluation datasets. In the following, we will describe two possible synthesis leakage cases, that are synthesis leakage due to oversampling and undersampling functions respectively. Oversampling methods are usually adopted to increase the dimension of the dataset when the available data instances are not enough, or the dataset is unbalanced in terms of data instances for each class. Example of oversampling strategies are SMOTE \citep{chawla2002smote} and ADASYN \citep{he2008adasyn}. 
A potential leakage risk arises when data instances used to evaluate the final model are also exploited by the oversampling method to generate the data. 
Conversely, undersampling methods reduce the number of instances in case of over-represented class(es) to create a more balanced distribution. There are different strategies for undersampling, such as Random undersampling \citep{prusa2015using}, which randomly removes instances from the majority class, and more sophisticated methods like Tomek Links \citep{tomek1976two} or Cluster Centroids \citep{yen2009cluster}, which selectively remove instances based on their proximity to other instances. Being methods based on proximity to other instances, the risk is that the removal is made relying on out-of-the-training data. }

\casestudybox{A Synthesis leakage due to oversampling}{
Consider a highly imbalanced dataset for a binary classification problem. Before adopting the ML pipeline, SMOTE oversampling algorithm is adopted. In a nutshell, SMOTE first selects a minority class instance at random and finds its $K$ nearest neighbors, with $K$ decided by the user. Then, a synthetic instance is generated  by performing linear interpolation  between the instance and one of the $K$ neighbors chosen at random. In such cases, there's a risk that the newly generated data points might be too similar or replicate data points which will be used to evaluate the classification model. In other words, the risk is that the chosen neighbors used to generate data will be placed in an evaluation set, such as $D^E$. Furthermore, over-sampled data can be wrongly incorporated into an evaluation set, evaluating the model also on synthetic but non effective data. \newtexttwo{An example of this scenario is reported in Fig. \ref{fig:oversampling}}.
\begin{figure}
    \centering
\includegraphics[width=0.8\linewidth]{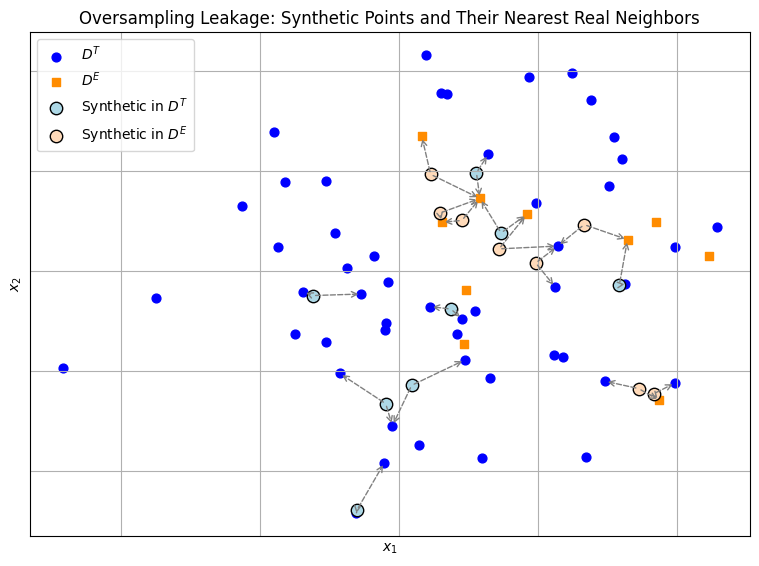}
    \caption{\newtexttwo{An example of an oversampled dataset generated exploiting data instances $D^E$  used to evaluate the trained model. Arrows indicate the real instances used to generate each synthetic data.}}
    \label{fig:oversampling}
\end{figure}
}

 

\casestudybox{A synthesis leakage due to undersampling}{Consider a highly imbalanced dataset for a binary classification problem. Before adopting the ML pipeline, Cluster Centroids undersampling method is adopted. Cluster Centroids replaces a cluster of data instances belonging to an over-represented class by its cluster centroid. In such case, if these centroids are computed using also data instances used for evaluation and subsequently incorporated into the training set, it's possible for the training process to exploit information similar to that of the evaluation set.
Furthermore, centroids can be wrongly incorporated into an evaluation set, evaluating the model also on synthetic but non effective data.}

\subsubsection{Data-induced leakage in Transductive Machine Learning and Transfer Learning}
In the previous subsection we considered data-induced leakage in classical ML. Now we analyze this kind of leakage in transductive ML and TL cases.
Since The aim of transductive ML isn't toward generalizing to unseen data beyond the provided dataset, synthesis leakage scenarios such as oversampling/undersampling without considering which data will be used in the evaluation can be considered safe in transductive cases. \newtexttwo{However,  direct and indirect label leakage remain possible and problematic. At the end of this section a possible scenario of data leakage due to labels in transductive ML is described.}

Conversely, in TL cases data-induced leakage may be easier to occur. For instance, in Domain Adaptation, data induced leakage can happen when possible relations between source domain data ${D}_{source}$ and target domain data ${D}_{target}$ are not considered in the experimental setup. For example, when data collected in different time windows are considered as different domains, but data acquired at a given time helps to infer the data acquired in other time windows in a way not foreseen by the task definition. 

Another issue arises when there is a lack of understanding of the data in the dataset construction process, in particular in Supervised DA cases. For example, a pre-trained model can be inadvertently fine-tuned on data related to the training set originally used to pre-train the model. In case of Domain Generalization, the risks involved are similar to the DA case, i.e., hidden relations not allowed in the task at hand between different domains can exist. 

The last two scenarios of this section describe cases of data-induced leakage in DA: the former involves data collection leakage due to data domains not properly accounted; the latter due to overlooking how the provided dataset was composed.

\casestudybox{A data collecting leakage in Transductive ML due to labels}{\newtexttwo{Suppose a library wants to classify its books by genre using ML. Each book is described by metadata fields such as title, author, ISBN, publication year, and shelf location. The task is transductive: all books to be classified are already available (the model does not need to generalize to future acquisitions), but the true genres of some books are initially unknown and must be predicted.
Among the data, the "shelf" feature results to be related to the book's genre. For example, mystery books are always shelved on the “MYS-” shelves, while science fiction books are on “SCI-” shelves. If this field is used as a feature, the model can trivially learn the mapping between shelf code and genre.}}

\casestudybox{A data collecting leakage in DA due to the different data domains}{Suppose we have a dataset of EEG data corresponding to different cognitive states (e.g., "focused" vs. "distracted"). The data are collected from a single subject across multiple sessions in a day. We want a ML model able to solve this classification task on a same subject (intra-subject). Different sessions are considered as belonging to different domains. The task goal is to classify cognitive states in a target domain using knowledge from the remaining source domains.
However, the experimental setup does not account for the fact that data acquired in nearby time windows may share strong correlations. For instance, EEG signals from the morning session might contain specific artifacts (e.g., eye blinks or muscle movements) that persist into the afternoon session. If these sessions are close in time, the model might inadvertently learn to rely on these artifacts, which are not truly reflective of generalizable cognitive states of the subjects. If the model is evaluated on data from a target session too close to the source ones, there is the risk that the model could exploit artifacts between nearby time windows. 
}
More in general, this scenario can occur when the time windows are too close between them, and artifacts present in the source time window can help the ML model in classifying the target time window. This can be allowed if the task requires predictions on nearby time windows, but conversely it should not be allowed if the task requires classification between time windows far from each other. 

\casestudybox{A data collecting leakage in DA due to data aggregation}{Suppose we are working on a ML model pretrained on a dataset $D_{I}$. To fine-tune the model, data from a Frankenstein dataset $D_{F}$ composed of data from several different sources are used. 
For fine-tuning, we select a subset of $D_{F}^{T} \subset D_{F}$ for training and the remaining $D_{F}^{E}=D_{F}\setminus D_{F}^{T}$ for evaluation, which is supposed to contain unseen data.
However, without considering the composition of $D_{F}$, it is possible that it contains also instances from $D_{I}$. Therefore, some of the data in $D_{F}^{E}$ can overlap with the original dataset $D_{I}$, i.e. ${D}_{F}^E \cap {D}_{I} \neq \emptyset$. Then, the model has already seen some of the evaluation data during the pre-training phase. Moreover, if over-sampling or under-sampling is performed on the entire ${D}_{F}$ without considering its composition, this behavior could be exacerbated.}

\subsubsection{Data-induced leakage: Discussion}
\label{sec:dataleaktypes:inner:discussion}
The previous scenarios illustrated how data leakage can occur as early as the data acquisition stage. In particular, the widespread use of publicly available datasets raises important questions about how these data have been aggregated. To highlight the potential impact of such leakage, we simulate a machine learning task involving Frankenstein datasets and evaluate the resulting performance. In particular, we designed an experiment where a model undergoes progressive training on a Frankenstein dataset composed  of multiple datasets, each potentially reusing portions of previously seen dataset.

We first generate an initial dataset $D_0$ for a binary classification problem. Then, for each subsequent stage $r \in \{1, \dots, R\}$, we construct a new dataset $D_r$ that includes a set of newly sampled instances along examples duplicated from one or more of the previously used datasets $\{D_0, \dots, D_{r-1}\}$. This mimics a Frankenstein dataset construction process, where data is collected or combined from multiple sources without explicit awareness of overlap or duplication.

At each stage, the classifier is trained on the cumulative data up to $D_r$, and its performance is evaluated using $5$-fold cross-validation over $D_r$. We compare two conditions:

\textit{With leakage:} training and test folds may contain overlapping examples due to duplication across datasets.
\textit{Without leakage:} we explicitly remove from the test fold any examples that are also present in the training fold (exact duplicates), ensuring a leakage-free evaluation.

This process is repeated 10 times to account for variability. Performance was evaluated using a DNN neural network with two hidden layers of 100 and 50 units respectively, ReLU activation functions. In Fig. \ref{fig:exp:data_collecting_leakage} We report the mean and standard deviation of accuracy for each stage, demonstrating how unnoticed reuse of previously seen data can lead to progressively inflated performance estimates as more Frankenstein datasets are added.

\begin{figure}
    \centering

\begin{tikzpicture}

\definecolor{darkgray176}{RGB}{176,176,176}

\begin{axis}[
tick align=outside,
tick pos=left,
title={Effect of Frankenstein Dataset Leakage on model performance},
x grid style={darkgray176},
xlabel={Dataset},
xmajorgrids,
xmin=0.85, xmax=9,
xtick style={color=black},
xtick={1,3,5,8},
xticklabels={
  \(\scriptscriptstyle   D_1\),
  \(\scriptscriptstyle   D_1\) \(\scriptscriptstyle \cup\) \(\scriptscriptstyle  D_2\),
  \(\scriptscriptstyle   D_1\) \(\scriptscriptstyle \cup\) \(\scriptscriptstyle  D_2\) \(\scriptscriptstyle \cup\) \(\scriptscriptstyle  D_3\),
  \(\scriptscriptstyle  D_1\) \(\scriptscriptstyle \cup\) \(\scriptscriptstyle  D_2\) \(\scriptscriptstyle \cup\) \(\scriptscriptstyle  D_3\) \(\scriptscriptstyle \cup\) \(\scriptscriptstyle  D_4\)
},
y grid style={darkgray176},
ylabel={Test Accuracy},
ymajorgrids,
ymin=0.683111428371253, ymax=0.932665973776496,
ytick style={color=black}
]
\path [draw=red, semithick]
(axis cs:1,0.884944081317682)
--(axis cs:1,0.921322585348985);

\path [draw=red, semithick]
(axis cs:3,0.832509686798064)
--(axis cs:3,0.876890313201936);

\path [draw=red, semithick]
(axis cs:5,0.80585837169459)
--(axis cs:5,0.84590162830541);

\path [draw=red, semithick]
(axis cs:8,0.7746422721814)
--(axis cs:8,0.813291061151934);

\addplot [semithick, red, mark=-, mark size=5, mark options={solid}, only marks]
table {%
1 0.884944081317682
3 0.832509686798064
5 0.80585837169459
8 0.7746422721814
};
\addplot [semithick, red, mark=-, mark size=5, mark options={solid}, only marks]
table {%
1 0.921322585348985
3 0.876890313201936
5 0.84590162830541
8 0.813291061151934
};
\path [draw=blue, semithick]
(axis cs:1,0.762052909210491)
--(axis cs:1,0.827524304716767);

\path [draw=blue, semithick]
(axis cs:3,0.726548929385618)
--(axis cs:3,0.800109778648626);

\path [draw=blue, semithick]
(axis cs:5,0.714749642611127)
--(axis cs:5,0.771794254800174);

\path [draw=blue, semithick]
(axis cs:8,0.694454816798764)
--(axis cs:8,0.748188934289239);

\addplot [semithick, blue, mark=-, mark size=5, mark options={solid}, only marks]
table {%
1 0.762052909210491
3 0.726548929385618
5 0.714749642611127
8 0.694454816798764
};
\addplot [semithick, blue, mark=-, mark size=5, mark options={solid}, only marks]
table {%
1 0.827524304716767
3 0.800109778648626
5 0.771794254800174
8 0.748188934289239
};
\addplot [semithick, red, mark=o, mark size=3, mark options={solid,fill opacity=0}, only marks]
table {%
1 0.903133333333333
3 0.8547
5 0.82588
8 0.793966666666667
};
\addplot [semithick, blue, mark=x, mark size=3, mark options={solid}, only marks]
table {%
1 0.794788606963629
3 0.763329354017122
5 0.743271948705651
8 0.721321875544002
};
\end{axis}

\end{tikzpicture}

\caption{\newtexttwo{Performance trends across Frankenstein datasets. The x-axis represents successive datasets constructed by combining newly generated data with portions of previously used datasets. The y-axis shows the resulting classification accuracy. Blue markers indicate performance under a clean setup (no data leakage), while red markers represent performance when data leakage is present due to repeated inclusion of previous data instances.}}
    \label{fig:exp:data_collecting_leakage}
\end{figure}
\color{black}

The data leakage case reported in \citep{li2020perils} and summarized in Sec. \ref{sec:dataleaktypes:inner:collect} about the inspected EEG classification task shows the importance to well design an experiment since the data collection. For instance, the acquisition of EEG signals belonging to the same class are gathered ``in block" from individual subjects. Each block, in essence, represents a distinct source of information. Consequently, when conducting analyses within a specific block, the model might exhibit relatively higher performance due to its familiarity with the data structure, patterns, and/or artifacts within that particular source. Moreover, this form of leakage poses challenges that persist even during subsequent steps, such as data splitting. Despite employing a well-thought-out split strategy between training and testing sets, mitigating the impact of potential artifacts in the data acquisition process isn't guaranteed. The presence of these artifacts might persist, affecting both the training and testing subsets, and potentially undermining the effectiveness of the data split, leading to other type of data leakage (see Sec. \ref{sec:dataleaktypes:splitrelated}). \newtexttwo{In \citep{rosenblatt2024data} is highlighted that family-related artifacts can occur in neuroimaging datasets due to oversampling of related individuals. Because brain structure and function are heritable, placing one family member in the training set and another in the test set can introduce leakage. Although family leakage had little to no effect on most prediction tasks, it slightly inflated performance in predicting attention problems.}
 
Incorporating the label as features arises from using data that ideally shouldn't be legitimately accessible to the model. Hence, preventing or circumventing leakage is linked to a clear definition of the task addressed and of a careful choice of the data inputs allowed for the predictive framework. 
A schematic representation of this type of leakage and a trivial solution is reported in Fig. \ref{fig:target_leakage}. In particular, in Fig. \ref{fig:target_leakage}a, each data instance $\vec{x}^{(i)}$ has the $f$-th feature equals to the label $y^{(i)}$. A possible solution is cleaning the data instances removing the $f$-th features, considering for each data instance only the $x^{(i)}_1,x^{(i)}_2,\dots x^{(i)}_{f-1}, x^{(i)}_{f+1}, \dots x^{(i)}_{d}$ features, as described in Fig. \ref{fig:target_leakage}b. However, including the $f$-th in the predictive model might be allowed or not, depending on the intended scope of the predictor.
\begin{figure}
\input{_IMG_CODE_TARGET_LEAK}
\label{fig:target_leakage}
\end{figure}
The authors of \citep{kapoor2023leakage} replicated an experimental assessment featured in a literature study influenced by this type of leakage. Their findings indicate that without this form of leakage
the results differ from the claims made by the original authors. 
To further assess the effect of this kind of indirect data leakage, we designed a synthetic binary classification task in which one feature, although semantically unrelated to the target, becomes spurious due to an artificial correlation introduced during data generation.

We constructed datasets with an additional feature whose statistical distribution was conditioned on the class label only in the leakage scenario. Specifically, this additional feature was sampled from class-dependent distributions with increasing separation, controlled by a parameter $\delta$ that regulates the leakage strength. Specifically, the feature simulates a variable whose distribution is artificially shifted based on the class: greater shifts ($\delta$) increase the correlation between the feature and the label, thus intensifying the leakage. In the non-leaky configuration, the same feature was sampled independently of the label, ensuring no spurious association.
For each value of $\delta$, we trained a simple DNN model with two layers of 100 and 50 neurons respectively and evaluated its accuracy on a hold-out set. We repeated the experiment 10 times and report the mean and standard deviation of the accuracy. The clean condition ($\delta = 0$) serves as a baseline to isolate the contribution of the leakage effect.

The resulting accuracy values are summarized in Fig. \ref{fig:exp:label_leakage}, which illustrates how performance improves as the spurious correlation is strengthened.

\begin{figure}
    \centering
\begin{tikzpicture}

\definecolor{darkgray176}{RGB}{176,176,176}

\begin{axis}[
tick align=outside,
tick pos=left,
title={Accuracy vs Degree of Indirect Leakage (± std)},
x grid style={darkgray176},
xmajorgrids,
xmin=-0.25, xmax=5.25,
xtick style={color=black},
xtick={0,1,2,3,4,5},
xticklabels={
  No leakage,
  \(\displaystyle \delta\)=5,
  \(\displaystyle \delta\)=10,
  \(\displaystyle \delta\)=15,
  \(\displaystyle \delta\)=20,
  \(\displaystyle \delta\)=25
},
y grid style={darkgray176},
ylabel={Accuracy},
ymajorgrids,
ymin=0, ymax=1.1,
ytick style={color=black}
]
\path [draw=blue, semithick]
(axis cs:0,0.72457485442715)
--(axis cs:0,0.904314034461739);

\addplot [semithick, blue, mark=-, mark size=10, mark options={solid}, only marks]
table {%
0 0.72457485442715
};
\addplot [semithick, blue, mark=-, mark size=10, mark options={solid}, only marks]
table {%
0 0.904314034461739
};
\path [draw=red, semithick]
(axis cs:1,0.86349233436343)
--(axis cs:1,0.95650766563657);

\path [draw=red, semithick]
(axis cs:2,0.980287047348029)
--(axis cs:2,0.98860184154086);

\path [draw=red, semithick]
(axis cs:3,0.997317540486252)
--(axis cs:3,1.00046023729153);

\path [draw=red, semithick]
(axis cs:4,1)
--(axis cs:4,1);

\path [draw=red, semithick]
(axis cs:5,1)
--(axis cs:5,1);

\addplot [semithick, red, mark=-, mark size=10, mark options={solid}, only marks]
table {%
1 0.86349233436343
2 0.980287047348029
3 0.997317540486252
4 1
5 1
};
\addplot [semithick, red, mark=-, mark size=10, mark options={solid}, only marks]
table {%
1 0.95650766563657
2 0.98860184154086
3 1.00046023729153
4 1
5 1
};
\addplot [semithick, blue, mark=x, mark size=3, mark options={solid,fill opacity=0}, only marks]
table {%
0 0.814444444444444
};
\addplot [semithick, red, mark=o, mark size=3, mark options={solid}, only marks]
table {%
1 0.91
2 0.984444444444444
3 0.998888888888889
4 1
5 1
};
\end{axis}

\end{tikzpicture}
  \caption{\newtexttwo{effect of indirect leakage on performance. Classification accuracy is shown as a function of the degree of leakage introduced via a spurious feature correlated with the class label. The blue point represents the baseline accuracy in a leakage-free setting, where the spurious feature is independent of the class. Red points indicate the accuracy under increasing leakage (higher $\delta$ values). }}
    \label{fig:exp:label_leakage}
\end{figure}

 \color{black}

More in general, determining whether a feature constitutes leakage isn't always straightforward. Let's consider the scenario where a feature serves as an indicator for a specific subset within the population. The legitimacy of using this information as a feature relies on whether a data instance assignment to a subset was a consequence of their task outcome or not. For example, as discusses in \citep{rosset2010medical} about medical data, the patient ID provided as a feature is a case of leakage whether the ID assignment is intrinsically linked to their medical condition, and it is not leakage if it's incidental and doesn't directly influence the predicted outcome. Indeed, if the ID code is correlated with specific geographical or demographic locations with higher disease prevalence rates, its association with the task outcome might be regarded as incidental. Consequently, the relevance of the ID code to predict the outcome might not directly derive from the medical condition itself, but could be due to other factors. Therefore, including this ID information in the predictive model might be allowed or not, depending on the intended scope of the predictor. In other words, if the objective is to develop a predictor that exploits contextual information like geographic location, incorporating the ID may be justifiable, given its association with regional or demographic factors. Conversely, if the aim is to create a predictor solely reliant on patient-specific information (therefore excluding other factors such as geographic location), then using the ID should be avoided.
\citep{rosset2010medical} highlights that several scenarios involving this kind of label leakage have a temporal component, where certain information is available at the time of the prediction, while outcomes and consequences of actions taken only become known later. For example, the health condition of a patient is known before the prediction, while the disease and the treatment are known later. Using only information available up to the decision-making point ensuring that it is not affected by any subsequent update shrinks the risk of unintentional leakage from future knowledge (such as outcomes or consequences) of actions taken given the outcomes.
Real labels inferred by spurious input features can be viewed as instances of the Clever Hans effect \citep{lapuschkin2019unmasking}. In a nutshell, Clever Hans happens when the training ML model is actually exploiting spurious features related to the label (see sec. \ref{sec:background:notation}).
To avoid the Clever Hans effect, it's essential to examine the features used for training, ensuring that they capture relevant information pertaining to the task at hand.
Clever Hans effects can be reduced exploiting current eXplainable Artificial Intelligence (XAI) methods \cite{ribeiro2016should,apicella2019contrastive}, locating which features or factors the model is primarily relying on for its predictions. 
However, XAI methods usually act on trained models, therefore can be used to locate Clever Hans at the end of a ML pipeline, and not to prevent it.
 
Finally, if over-sampling or under-sampling is performed on the whole ${D}$ without distinguish between train and evaluation data, the risk is twofold: firstly, that information from the evaluation data is used to generate the training data set, favoring instances that are too similar to the evaluation data. Secondly, there's a risk that synthetic data generated will be included in the evaluation sets, potentially distorting the evaluation with synthetic data or ignoring outliers that may be important to the task at hand. To mitigate these risks, it is advisable to conduct any under-sampling or over-sampling procedures only after implementing the data split strategy, ensuring that these techniques are applied properly to the interested subsets. 

Fig. \ref{fig:oversampling} illustrates the performance impact of increasing overlap between the test set and the data used during oversampling with SMOTE in a binary classification task. Results are averaged over 10 experiments using 5-fold cross-validation on synthetic data. A simple two layer DNN with 100 and 50 neurons respectively was used for evaluation. Error bars indicate standard deviation. The x-axis represents the proportion of test samples included in the data used to perform SMOTE (i.e., overlap ratio). The y-axis shows the classification accuracy on the  test set. While SMOTE is intended to address class imbalance by synthetically generating new samples, including instances without distinguish between evaluation sets and training set in the oversampling process introduces a form of synthesis leakage.
As the overlap increases, the classifier’s accuracy improves, reflecting an inflated evaluation due to the leakage. 

\begin{figure}
\centering
\begin{tikzpicture}
\definecolor{darkgray176}{RGB}{176,176,176}

\begin{axis}[
tick align=outside,
tick pos=left,
title={Accuracy vs SMOTE-Test Overlap (K-Fold CV, ± std)},
x grid style={darkgray176},
xmajorgrids,
xmin=-0.3, xmax=6.3,
xtick style={color=black},
xtick={0,1,2,3,4,5,6},
xticklabels={\tiny No overlap, \tiny 20\%,\tiny 40\%,\tiny 50\%,\tiny 60\%,\tiny 80\%,\tiny 100\%},
y grid style={darkgray176},
ylabel={Accuracy},
ymajorgrids,
ymin=0, ymax=1,
ytick style={color=black}
]
\path [draw=blue, semithick]
(axis cs:0,0.736710471491259)
--(axis cs:0,0.834956195175408);

\addplot [semithick, blue, mark=-, mark size=10, mark options={solid}, only marks]
table {%
0 0.736710471491259
};
\addplot [semithick, blue, mark=-, mark size=10, mark options={solid}, only marks]
table {%
0 0.834956195175408
};
\path [draw=red, semithick]
(axis cs:1,0.753275981248464)
--(axis cs:1,0.849807352084869);

\path [draw=red, semithick]
(axis cs:2,0.77355167860501)
--(axis cs:2,0.863364988061657);

\path [draw=red, semithick]
(axis cs:3,0.785756618162354)
--(axis cs:3,0.870743381837646);

\path [draw=red, semithick]
(axis cs:4,0.793903660632464)
--(axis cs:4,0.878679672700869);

\path [draw=red, semithick]
(axis cs:5,0.815233864887074)
--(axis cs:5,0.892182801779593);

\path [draw=red, semithick]
(axis cs:6,0.833354564251591)
--(axis cs:6,0.904312102415075);

\addplot [semithick, red, mark=-, mark size=10, mark options={solid}, only marks]
table {%
1 0.753275981248464
2 0.77355167860501
3 0.785756618162354
4 0.793903660632464
5 0.815233864887074
6 0.833354564251591
};
\addplot [semithick, red, mark=-, mark size=10, mark options={solid}, only marks]
table {%
1 0.849807352084869
2 0.863364988061657
3 0.870743381837646
4 0.878679672700869
5 0.892182801779593
6 0.904312102415075
};
\addplot [semithick, blue, mark=x, mark size=3, mark options={solid,fill opacity=0}, only marks]
table {%
0 0.785833333333333
};
\addplot [semithick, red, mark=o, mark size=3, mark options={solid}, only marks]
table {%
1 0.801541666666667
2 0.818458333333333
3 0.82825
4 0.836291666666667
5 0.853708333333333
6 0.868833333333333
};
\end{axis}
\end{tikzpicture}
\caption{\newtexttwo{Performance impact of increasing overlap between the test set and the data used during oversampling with SMOTE in a binary classification task. The x-axis represents the proportion of test samples included in the data used to perform SMOTE (i.e., overlap ratio). The y-axis shows the classification accuracy on the  test set. The blue point represents the baseline accuracy in a leakage-free setting, where the spurious feature is independent of the class. Red points indicate the accuracy under increasing leakage.}}
\label{fig:exp:oversampling}
\end{figure}
\color{black}

\newtexttwo{Similar conclusions are drawn in \cite{whalen2022navigating} for the genomics domain, where class balancing methods are explicitly recommended to be applied strictly within the training fold..
In other words, it can be a good rule of thumb to over-sample/under-sample only training data, letting untouched the evaluation one. }
A schematic representation of this type of leakage and solution is illustrated in Fig. \ref{code:synthesis_leakage}. \newtexttwo{In particular, in Fig. \ref{code:synthesis_leakage}a the $g$ function generates a new dataset $D_{gen}$ oversampling or undersampling the data instances relying on the input dataset $D$. The split between the training set $D^T$ and the evaluation set $D^E$ from the generated dataset $D_{gen}$ introduces the risk that $D^T$ may consist of data synthesized by information now incorporated into $D^E$, and vice versa. Consequently, the model's performance on $D^E$ might be overestimated due to the unintentional inclusion of information present in both $D^T$ and $D^E$ during the training phase. A preventive measure to mitigate this type of leakage involves synthesizing the data exclusively on $D^T$  after the data split, effectively avoiding the entanglement of training and evaluation data during the synthesis process, as shown in Fig.\ref{code:synthesis_leakage}b}.
\begin{figure}
\input{_IMG_CODE_SYNTHESIS_LEAKAGE}
\label{code:synthesis_leakage}
\end{figure}

\newtext{Considering transductive ML, oversampling or undersampling without regard to evaluation data is in general safe. Instead, data leakage is more likely in TL cases, particularly in DA and DG. Leakage can occur if relationships between source and target domain data are not properly accounted for, such as when data from different time periods are treated as separate domains but still influence each other, or when a pre-trained model may inadvertently be fine-tuned on data related to its original training set. A similar case is reported in \citep{wen2020convolutional} as a case of \textit{biased transfer learning}. However, we want to emphasize that in this scenario the bias stems as a consequence of data leakage due to the TL strategy adopted, rather than a bias within the data itself. If these datasets were used in separate, independent ML models, there wouldn't be any inherent bias present in the data.}

\subsection{\textbf{Preprocessing-related leakage}}
\label{sec:dataleaktypes:preproc}
We consider leakage before data split when the leak of information happens before the available dataset is divided into training and out-of-the-training data (see Fig. \ref{fig:leakage_type}). 
In particular, preprocessing leakage typically arises when preprocessing procedures (e.g., min-max scaling or normalization) are applied before a proper separation between training and evaluation data. These procedures require parameters (e.g., feature ranges or statistics) that are either predefined by the task or inferred from the data. Formally:
\begin{definition} \newtexttwo{let $f: \mathbb{D} \times \mathbb{Q} \to \mathbb{D}$ be a preprocessing function that transforms a dataset $D$ using parameters $Q \in \mathbb{Q}$, and let $\upsilon: \mathbb{D} \to \mathbb{Q}$ be a function estimating such parameters from data. If the model is trained on $D^T \subseteq D_{\text{prep}}$ and evaluated on $D^E \subseteq D_{\text{prep}}$, with $D_{\text{prep}} = f(D, Q)$ and $Q = \upsilon(D')$ for some $D' \supseteq D^E$, then there is a risk that training and evaluation data share information, introducing leakage.}
\end{definition}
In other words, the information shared between evaluation set and parameter estimation introduces task-spurious information into the data.
A common case occurs when $D' = D$, i.e., the preprocessing parameters are computed from the full dataset but leakage may still occur even when $D' \not= D$, depending on how $\upsilon$ exploits the available data.
Moreover, when the normalization function depends, explicitly or implicitly, on label information, such as supervised PCA \citep{bair2006prediction}, the risk is even greater. In these cases, the preprocessing procedure itself needs knowledge about the target variable, similarly to  label leakage discussed in Sec. \ref{sec:dataleaktypes:label}, but in a different stage of the pipeline.

 
Depending on the $f$ adopted, we can distinguish between:
\newtext{\begin{enumerate}[i.]
    \item \textit{Normalization leakage}: when the function $f$ is a normalization function;
    \item \textit{Cleaning leakage}: when the function $f$ is a cleaning function;
    \item \textit{Imputation leakage}: when the function $f$ is an imputation function;
    \item \textit{Feature engineering leakage}: when the function $f$ is a feature engineering function, such as a feature selection or a feature extraction technique.
\end{enumerate}}
\subsubsection{Normalization leakage}
\label{sec:dataleaktypes:preproc:norm}
Normalization leakage arises when a normalization function is applied to each data point in a training/out-of-training subset using parameters calculated over the entire dataset provided. 

For example, min-max scaling \citep{herwanto2021comparison} procedure projects each data point feature in a given [min, max] range. However, min-max scaling needs the actual feature ranges available. In scenarios where these ranges are known \textit{a priori} this is not a problem. For example, in image domain it is known that the pixel values typically fall within the bounded range of integer numbers in $[0,255]$. In other cases, instead, feature ranges have to be estimated from the available data. 
In the following, a scenario involving this kind of leakage is described.

\begin{figure}
    \centering
    \includegraphics[width=0.8\linewidth]{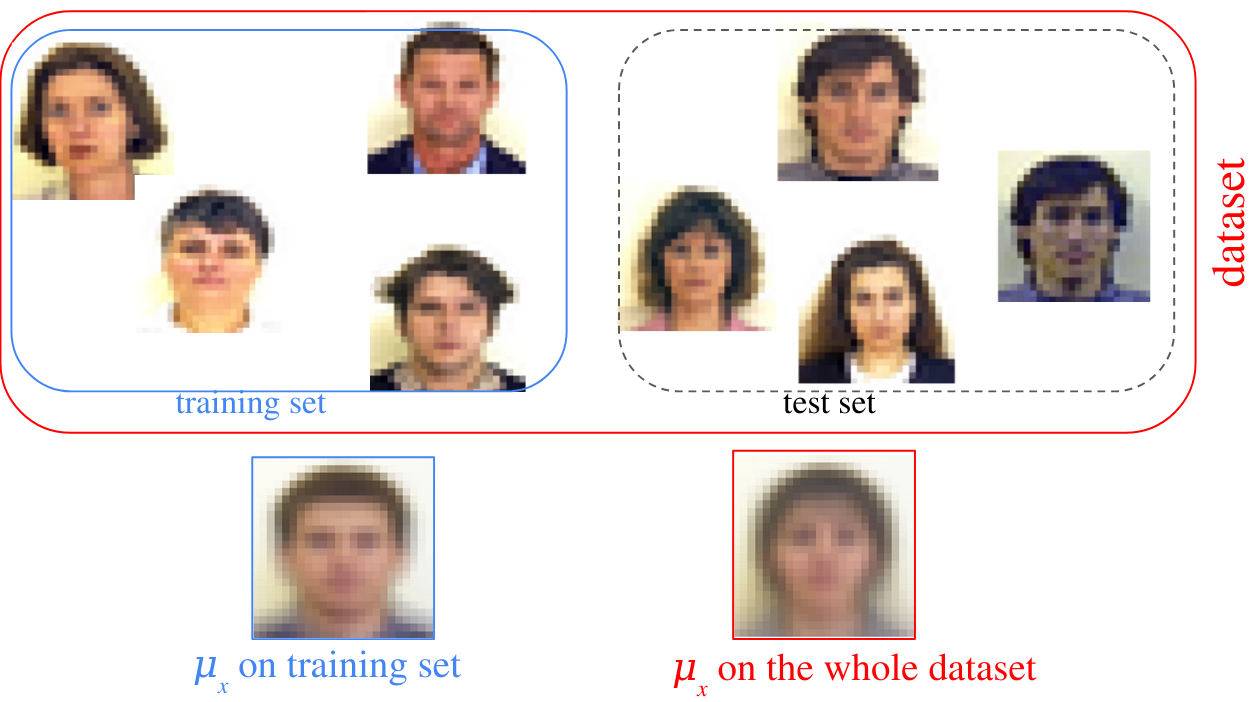}
    \caption{\newtexttwo{An example of preprocessing leakage due to wrong data used to compute standardization parameters. The figure shows that the average $\mu$ computed on the whole dataset is significantly different from the average computed on the training set. Therefore, using full dataset average to standardize the data provides information about the test data to the training data.}}
    \label{fig:standardization}
\end{figure}

\casestudybox{A normalization leakage due to min-max scaling}{Suppose that a dataset is composed of data with features $\vec{x}$. The data features have to be projected into the range $[0,1]$ by a min-max scaling function $f:\mathbb{D} \times \mathbb{Q} \to \mathbb{D}$\\ where $\mathbb{Q}$ is the of the function parameters need to make the transformation (in this case, $\vec{q} \in \mathbb{Q}$ is composed of the minimum and maximum values of each feature). Since the actual boundaries of the feature range are unknown, the scaling parameters $\vec{q}$ are inferred from the available dataset $D$. After normalization, a train-test split is made. Since the scaling parameters are computed on the whole dataset $D$, the resulting scaled training data $D^T$ incorporates information from the test set $D^E$. Indeed, although $D^E$ isn't used directly in the training process, knowledge about its distribution (through the min and max values in $\vec{q}$ used during the normalization) can still bias the model. }
Similarly, standardization \citep{herwanto2021comparison,apicella2023effects} maps each feature's data points to a normal distribution using mean and standard deviation estimated by the provided data instances.  
However, the function's parameters $\vec{q}$ (minimum and maximum for min-max scaling, mean and standard deviation for standardization) should be estimated only on the data available during training (i.e. $D^T$), without considering out-of-the-training data (such as $D^E$). Instead, they are often computed using all the data points provided. Using all the available data for estimating the preprocessing function parameters can impact the performance on evaluation data, potentially leading to an overestimation of the model due to unauthorized knowledge (given by the estimated parameters) available during the training stage coming from the out-of-the-training data. \newtexttwo{An example of this kind of leakage is shown\footnote{images in figure generated using data available to \url{https://pics.stir.ac.uk/}} in Fig. \ref{fig:standardization}.}
\newtexttwo{\cite{whalen2022navigating} identify this form of data leakage in genomics as leaky preprocessing, highlighting how it frequently occurs in the literature. They emphasize that many common transformations, such as standardization, PCA, and other common preprocessing methods, can unintentionally introduce information from the test set into the training process, thereby compromising the validity of model evaluation.} 
Obviously, preprocessed data are not affected by this kind of data leakage if the function's parameters  $\vec{q}$ are not computed on the data but are already available by the domain or are given by the task. 

\subsubsection{Cleaning leakage}
\label{sec:dataleaktypes:preproc:clean}
Data instances may contain various types of noise that can negatively impact model training. To mitigate this, proper data cleaning techniques are typically employed. However, if not carefully handled, these cleaning processes can inadvertently introduce data leakage, compromising the integrity of the model. In the following of this section a possible scenario is described.

\casestudybox{A cleaning leakage due to moving average}{\cite{kuhn2019feature} supposed that a dataset is composed of several sequential data (e.g., a time series). To reduce noise, the data features are preprocessed with a moving average where the original feature values are replaced with smoothed values. After smoothing, a train-test split is made. Since the data is sequential, leakage can occur if the last data point in the training set uses the first data point of the evaluation set to compute the moving average that replaces its values. In this scenario, the final training data incorporates information from the test set, leading to biased model training.}
In the case described above, the resulting leakage can be considered minimal since only the first test data points are involved in the normalization procedure. However, more severe and serious cases may occur. 
In general, determining the optimal cleaning strategy depends on factors such as the specific task at hand, its domain and the available data, together with the ML algorithm utilized.
Furthermore, it is important to consider if the cleaning methods require knowledge from several data instances or not. Indeed, 
cleaning functions can need parameters computed on several instances, arising data leakage scenarios similar to normalization leakage .

\subsubsection{Imputation leakage} 
\label{sec:dataleaktypes:preproc:imput}
In many scenarios, feature vectors $\vec{x}$ may have missing values for one or more components due to several reasons, such as human error during data collection or wrong acquisitions. Usually, imputation is made estimating the missing values based on the available data, for example substituting the missing value with the mean or median \citep{zhang2016missing} of the existing ones, or with more complex methods such as clustering \citep{zhang2008missing} or KNN \citep{troyanskaya2001missing}. However, also in this case if the missing value computation is made among the whole dataset, leakage can happen.
In the following of this section a possible scenario of imputation leakage is described. In \citep{kapoor2023leakage} at least three studies affected by this type of leakage were identified.

\casestudybox{An imputation leakage due to neighbor data}{Suppose a dataset contains complete and incomplete instances, i.e. the feature value $x^{(u)}_i$ is missing for a subset of $U$ instances $\{\vec{x}^{(u)}\}_{u=1}^{U}\subset D$. To handle this, the KNNimpute algorithm \citep{troyanskaya2001missing} is applied to estimate the missing features. For each instance $\vec{x}^{(u)}$, KNNimpute computes the missing value as a weighted average of the $K$ nearest data instances that have a value present for $x_i$. 
Once imputation is complete, the dataset is split into training and test sets. However, because the missing feature values are estimated using $K$ neighboring data instances, it is possible that one or more instances in the training set were imputed using neighbors from an evaluation set. Conversely, the evaluation set may also have imputed values calculated using neighbors from the training set \newtexttwo{(see Fig. \ref{fig:imputation} for an example)}. As a result, the training data may incorporate information from the test set and vice versa, leading to biased model training and evaluation. Although the test data isn't used directly in training, the imputation process involving test data can still bias the model. 
\begin{figure}
    \centering
\includegraphics[width=0.8\linewidth]{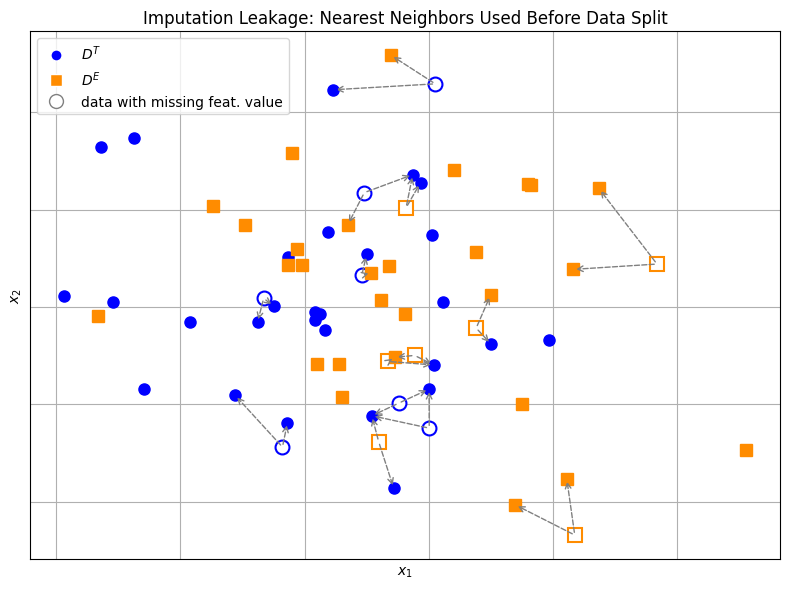}

    \caption{\newtexttwo{An example of imputation leakage. To generate the missing feature values, data exploits knowledge both from data that will be assigned to the training $D^T$ and the evaluation $D^E$ set.}}
    \label{fig:imputation}
\end{figure}
}

\subsubsection{Feature engineering leakage} 
\label{sec:dataleaktypes:preproc:feat}
Some feature engineering strategies such as feature selection methods remove the less relevant features from the original dataset reducing computational complexity and the risk of over-fitting. 
However, performing a feature selection strategy on the whole dataset risks selecting discriminative features based on data used in the evaluation step, resulting in a leakage scenario similar to the normalization leakage. \citep{shim2021inflated} compares the classification performance between feature selection performed on all the available data and features selected letting out the evaluation instances. The differences between the performance can be viewed as a form of over-fitting due to information leakage, indicating the importance of correctly use feature selection strategies.

In the following illustrative scenario we summarize a literature case of this type of leakage.

\casestudybox{A feature engineering leakage due to feature selection}{
In \citep{zhdanov2020use}, two ML methods are applied in prognosis of depression treatment with escitalopram using EEG data. A feature selection strategy based on $p$-values from an unpaired t-test is used to identify five feature sets. For each feature set, a cross-validation strategy is employed to create training and out-of-training sets for model training and evaluation. However, since the feature selection is performed before splitting the data, the selection process is influenced by the entire dataset, potentially biasing both the model and its evaluations.
}
\newtext{\citep{pulini2019classification} made an analysis on $69$ studies for the Attention-Deficit/Hyperactivity Disorder (ADHD) classification, detecting that at least the $15.9 \%$ of the inspected studies are affected by this type of leakage
due to applying feature selection procedure to the whole datasets without considering data split.}

\subsubsection{Preprocessing-related leakage: a particular case}
\label{sec:preprop:particular}
\newtext{In this subsection, we want to emphasize how this type of leakage can be insidious and difficult to detect. We will present a scenario where this leakage subtly affects the data, making it harder to recognize. In particular, we will discuss how this kind leakage can arise adopting some common ML training algorithms,such as Gradient Descent with early-stopping criteria on ANNs/DNNs, that use a \textit{validation} dataset $D^{V}$ to monitor model performance during training.
}

\casestudybox{A pre-processing leakage affecting the training step leading to underestimated performance}{Suppose we have a dataset splitted into $D^{T}$ and $D^{E}$. We want to training an ANN model on $D^{T}$ using Gradient Descent, where the model iteratively updates its parameters in multiple steps. The training process is monitored using a validation set $D^{V}$ to implement early stopping. A common practice is get $D^{V}$ from $D^{T}$ splitting it into two parts: an effective training subset $D^{T'}$ and a validation subset $D^{V}$ (i.e., $D^T = D^{T'} \cup D^V$ and $D^{T'} \cap D^{V} = \emptyset$).
However, before performing the split between $D^V$ and $D^{T'}$, we apply data preprocessing steps, such as standardization, on the whole $D^{T}$. To make this, we compute the mean and standard deviation of feature values using  $D^{T}$ and apply these statistics for normalize it and $D^{E}$.
Although $D^{V}$ is kept separate from $D^{T'}$ during the training, the validation set indirectly influences the training process since the model has access to statistics derived from the whole $D^{T}= D^{T'} \cup D^V$.
While this does not represent leakage between the training and the final evaluation set $D^{E}$, it compromises the integrity of the model selection process. Since the model has been tuned using information and statistics from both $D^{T'}$ and $D^{V}$, the model selected by the early-stopping based on $D^{V}$ might not be the best model when tested on unseen data $D^{E}$, leading to overfitting or under-training.} \newtexttwo{This situation is represented in Fig. \ref{fig:validation_leak}: on the left, the training set $D^{T'}$ and the validation set $D^V$ are impacted by data leakage described in Fig. \ref{code:validation_leak}a. Consequently, the model selected minimizes loss on $D^V$, potentially not accurately reflecting its performance on the out-of-the-training data represented by $D^E$. On the right, if $D^{T'}$ is unaffected by data leakage described in Fig. \ref{code:validation_leak}b, $D^V$ is a more representative sample of out-of-the-training data, including $D^E$. As a result, the model selected exhibits superior generalization performance on $D^E$.}
 \begin{figure}
\input{_IMG_VALIDATION_LEAK}
\label{fig:validation_leak}
\end{figure}

\subsubsection{Preprocessing-related leakage in Transductive Machine Learning and Transfer Learning} 
\label{sec:dataleaktypes:preproc:trans}
In transductive machine learning, all available  input instances $\vec{x}$ are accessible during training, including those from the evaluation set. As a result, operations that would typically cause normalization or feature engineering leakage in inductive settings, such as computing normalization parameters or selecting features based on the entire dataset—are generally acceptable in the transductive framework. However, leakage can still occur when information from the evaluation labels, which are supposed to remain unseen, is used during preprocessing. 

In TL, instead, the main aim is to transfer knowledge learned from one task or domain to another, related but different, task or domain. 
However, In several TL cases, it should be also considered if the aim of the task is to generalize to new unseen data (inductive case) or if we want to solve our task only on known available unlabeled data from a different domain/task (transductive case).
Considering the classical TL subfamilies, DA strategies usually assumes that an unlabeled dataset ${D}_{target}$ from a target domain is available during the training stage, together with a labeled dataset ${D}_{source}$ from a source domain. In this case, the desired aim can be one of the following: i) we are interested in resolving a task only on the provided ${D}_{target}$ (similarly to the transductive ML case), or ii) we want a model able to generalize to the whole domain from which  ${D}_{target}$ has been acquired. In the former, since the target is just to correctly resolve the task on the specific ${D}_{target}$ dataset, we can consider safe to use all the available data (both from the source and the target domain) to train the classifier, and also to compute pre-processing parameters or executing feature engineering procedures. Conversely, if the aim is to generalize toward new unseen data coming from the target domain, the model evaluation should be made on target domain data not used to compute normalization parameters or to make feature engineering operations such as feature selection. In case of DG, we assume that several dataset $\{D_{source_s}\}_{s=1}^S$ from $S>1$ source domains are available, and no data on the target domain is known during the training. 
Indeed, ${D}_{target}$ is usually available only for evaluation purposes on the target domain. As a consequence, the risks associated with leakage before the data splitting stage remain similar to those encountered in standard ML paradigm. For instance,  computing normalization parameters using information not only from the available sources, but also from the target data, can impact the model's performance evaluation.

Three possible scenarios of preprocessing-related leakage in Transductive ML and DA by are described in the following.

\casestudybox{A Preprocessing-related leakage in Transductive ML}{
\newtexttwo{A literary research team aims to attribute a fixed set of anonymous texts to a group of known authors. Some documents are labeled with their true authorship, and the remaining documents are unlabeled and must be classified. The team is interested in predicting only the labels of the current set of documents, not generalize to future ones. To reduce the dimensionality of the feature space (e.g., stylometric measures, word embeddings), the team applies Supervised PCA (SPCA,\cite{bair2006prediction}) to the full labeled dataset, without considering that a part of them $D^E$ has to be used after learning for evaluation. As a result, the training data $D^T$ are pre-proprocessed exploiting knowledge about the labels of $D^E$, making the final evaluation unreliable.}
}

\casestudybox{A Preprocessing-related leakage in DA due to an EEG normalization method}{\newtexttwo{A common setting in domain adaptation for EEG classification involves training a model on a labeled source domain and evaluating it on a different, unlabeled target domain. A popular preprocessing method in this context is Common Spatial Pattern (CSP, \citep{koles1990spatial}), which relies on estimating class-specific covariance matrices to maximize the variance difference between classes.
A source of data leakage in this scenario arises when CSP filters are estimated using data from both the source and the target domain, even if the labels from the target domain are not used. This practice may appear innocuous, particularly when the target domain is included only to "stabilize" the covariance estimation. However, since CSP optimizes a transformation that is sensitive to the data distribution, incorporating the target data, even in an unsupervised way—can introduce information about the evaluation set into the training process, thus biasing the learned filters and inflating performance.}}

\casestudybox{A preprocessing-related data leakage in DA due to data normalization}{Suppose we are working on a DA task where the goal is to classify emotions using EEG data from new, unseen subjects. We have EEG recordings from a single subject across multiple sessions, representing the source domain dataset $D_{source}$, and EEG recordings from an entirely different subject, representing the target domain dataset $D_{target}$. The goal is to train a model on $D_{source}$,  and adapt it to generalize well to unseen data  $D_{target}$ from the target domain. Since a DA strategy is adopted, the ML model can exploit both train and target data feature vector during the training, i.e. $D_{source}$ in supervised way and $D_{target}$ in unsupervised way.
During data preparation, a standardization procedure $f:\mathbb{D} \times \mathbb{Q} \to \mathbb{D}$ where $\mathbb{Q}$ \text{is the space of the parameters of $f$} is applied across both the source and target domain datasets. For instance, the mean and standard deviation composing the function parameters $\vec{q} \in \mathbb{Q}$ for each feature are computed using all the available data, including data from the target domain.
As a result, the model indirectly benefits from information it wouldn't have access to in a real-world scenario, leading to the overly optimistic results that don't reflect true generalization to new data.} 

\subsubsection{Preprocessing-related data leakage: Discussion}
\label{sec:dataleaktypes:preproc:general_disc}
In all the above discussed case, the model may become overly specialized to perform well on the out-of-the-training data, such as evaluation set(s). Consequently, this might not accurately reflect its generalization ability to handle new, unseen data outside the evaluation set(s). 
To evaluate the impact of normalization leakage, we designed a synthetic classification task in which a shift is artificially introduced in the test data distribution, and compared two different normalization strategies.

We generated multiple datasets using a standard binary classification benchmark. In each experiment, we applied a shift to all features in the test set, simulating a distributional discrepancy between training and testing data. 

We repeated the experiment 10 times and used 5-fold stratified cross-validation to assess performance on a simple DNN model with two layers of 100 and 50 neurons.

The results are summarized in Figure \ref{fig:exp:normalization_leakage}. As the shift magnitude increases, performance in the leakage-free (blue) condition differs from the leakage-affected condition (red), reflecting the model's inability to adapt to the distributional change. In contrast, the leaky normalization compensates for the shift by exploiting information from the test set during preprocessing, resulting in artificially stable performance across all conditions.

It is worth noting that, under ideal conditions where the available sample is highly representative of the overall data distribution, the statistics used for normalization (e.g., mean and standard deviation) computed on the training set may closely resemble those of the whole dataset. In such cases, the impact of normalization leakage might appear negligible. However, this assumption rarely holds in real-world scenarios, where sample biases, domain shifts, or limited data availability may cause significant discrepancies. Therefore, it is essential to estimate normalization parameters exclusively from the training data to ensure a realistic and unbiased evaluation.

\begin{figure}
    \centering
\begin{tikzpicture}

\definecolor{darkgray176}{RGB}{176,176,176}

\begin{axis}[
tick align=outside,
tick pos=left,
title={Normalization Leakage vs Shift Magnitude},
x grid style={darkgray176},
xlabel={Shift added to test set},
xmajorgrids,
xmin=-0.5, xmax=10.5,
xtick style={color=black,rotate=0},
xtick={0,1,2,3,4,5,6,7,8,9,10},
xticklabel style={rotate=90},
xticklabels={
  \tiny shift=0.0,
  \tiny shift=0.5,
  \tiny shift=1,
  \tiny shift=1.5,
  \tiny shift=2,
  \tiny shift=2.5,
  \tiny shift=3.0,
  \tiny shift=3.5,
  \tiny shift=4.0,
  \tiny shift=4.5,
  \tiny shift=5.0
},
y grid style={darkgray176},
ylabel={Accuracy},
ymajorgrids,
ymin=0, ymax=1.05,
ytick style={color=black}
]
\path [draw=red, semithick]
(axis cs:0,0.874723644434197)
--(axis cs:0,0.966476355565803);

\path [draw=red, semithick]
(axis cs:1,0.777140130911148)
--(axis cs:1,0.926859869088851);

\path [draw=red, semithick]
(axis cs:2,0.659156315032217)
--(axis cs:2,0.919243684967783);

\path [draw=red, semithick]
(axis cs:3,0.610485598752286)
--(axis cs:3,0.893114401247714);

\path [draw=red, semithick]
(axis cs:4,0.587514310453835)
--(axis cs:4,0.884485689546165);

\path [draw=red, semithick]
(axis cs:5,0.565047190948015)
--(axis cs:5,0.884552809051985);

\path [draw=red, semithick]
(axis cs:6,0.546733687491546)
--(axis cs:6,0.860466312508454);

\path [draw=red, semithick]
(axis cs:7,0.52736028600639)
--(axis cs:7,0.84063971399361);

\path [draw=red, semithick]
(axis cs:8,0.512928475348405)
--(axis cs:8,0.816671524651595);

\path [draw=red, semithick]
(axis cs:9,0.501760562932544)
--(axis cs:9,0.803039437067456);

\path [draw=red, semithick]
(axis cs:10,0.498928291641014)
--(axis cs:10,0.807471708358986);

\addplot [semithick, red, mark=-, mark size=5, mark options={solid}, only marks]
table {%
0 0.874723644434197
1 0.777140130911148
2 0.659156315032217
3 0.610485598752286
4 0.587514310453835
5 0.565047190948015
6 0.546733687491546
7 0.52736028600639
8 0.512928475348405
9 0.501760562932544
10 0.498928291641014
};
\addplot [semithick, red, mark=-, mark size=5, mark options={solid}, only marks]
table {%
0 0.966476355565803
1 0.926859869088851
2 0.919243684967783
3 0.893114401247714
4 0.884485689546165
5 0.884552809051985
6 0.860466312508454
7 0.84063971399361
8 0.816671524651595
9 0.803039437067456
10 0.807471708358986
};
\path [draw=blue, semithick]
(axis cs:0,0.874723644434197)
--(axis cs:0,0.966476355565803);

\path [draw=blue, semithick]
(axis cs:1,0.76858338062809)
--(axis cs:1,0.915416619371911);

\path [draw=blue, semithick]
(axis cs:2,0.652950233470506)
--(axis cs:2,0.920649766529494);

\path [draw=blue, semithick]
(axis cs:3,0.631296436656779)
--(axis cs:3,0.89790356334322);

\path [draw=blue, semithick]
(axis cs:4,0.600292686065123)
--(axis cs:4,0.850507313934878);

\path [draw=blue, semithick]
(axis cs:5,0.568593150885746)
--(axis cs:5,0.802206849114253);

\path [draw=blue, semithick]
(axis cs:6,0.539492820513951)
--(axis cs:6,0.772107179486049);

\path [draw=blue, semithick]
(axis cs:7,0.518643260689824)
--(axis cs:7,0.749756739310176);

\path [draw=blue, semithick]
(axis cs:8,0.507288862700287)
--(axis cs:8,0.714711137299713);

\path [draw=blue, semithick]
(axis cs:9,0.492585448748782)
--(axis cs:9,0.707014551251218);

\path [draw=blue, semithick]
(axis cs:10,0.482952074940805)
--(axis cs:10,0.705847925059195);

\addplot [semithick, blue, mark=-, mark size=5, mark options={solid}, only marks]
table {%
0 0.874723644434197
1 0.76858338062809
2 0.652950233470506
3 0.631296436656779
4 0.600292686065123
5 0.568593150885746
6 0.539492820513951
7 0.518643260689824
8 0.507288862700287
9 0.492585448748782
10 0.482952074940805
};
\addplot [semithick, blue, mark=-, mark size=5, mark options={solid}, only marks]
table {%
0 0.966476355565803
1 0.915416619371911
2 0.920649766529494
3 0.89790356334322
4 0.850507313934878
5 0.802206849114253
6 0.772107179486049
7 0.749756739310176
8 0.714711137299713
9 0.707014551251218
10 0.705847925059195
};
\addplot [semithick, red, mark=o, mark size=3, mark options={solid}, only marks]
table {%
0 0.9206
1 0.852
2 0.7892
3 0.7518
4 0.736
5 0.7248
6 0.7036
7 0.684
8 0.6648
9 0.6524
10 0.6532
};
\addplot [semithick, blue, mark=x, mark size=3, mark options={solid,fill opacity=0}, only marks]
table {%
0 0.9206
1 0.842
2 0.7868
3 0.7646
4 0.7254
5 0.6854
6 0.6558
7 0.6342
8 0.611
9 0.5998
10 0.5944
};
\end{axis}

\end{tikzpicture}

\caption{\newtexttwo{Impact of normalization leakage under increasing test shift. As the shift magnitude increases, performance in the leakage-free (blue) condition differs from the leakage-affected condition (red).}}
\label{fig:exp:normalization_leakage}
\end{figure}
\color{black}
However, the aforementioned leakage instances can be mitigated by dividing the data into separate training and out-of-the-training sets beforehand, as described in Fig. \ref{code:preprocessing_leakage}. \newtexttwo{We consider a preprocessing function $f$ working on a dataset relying on a set of further parameters $Q$. As an example, let $f(D, Q)$ represent the application of the classical standardization function $\frac{\vec{x}^{(i)}-\mu}{\sigma}$ on all the data instances of $D$, where $\mu$ and $\sigma$ are parameters provided by $Q = {\mu, \sigma}$. Each element $\vec{x}^{(i)}$ in the dataset $D$ is standardized using the mean $\mu$ and standard deviation $\sigma$ specified in the parameter set $Q$. In Fig. \ref{code:preprocessing_leakage}a, the parameter set $Q$ is computed by the function $\upsilon$ on the entire dataset $D$. Consequently, all data instances in the preprocessed dataset $D_{prep}$ are influenced by information computed across the whole $D$. This information, retained in both the training set $D^T$ and the evaluation set $D^E$, has the potential to bias the training stage. Although the resulting model may exhibit superior performance on $D^E$, this performance might not accurately reflect the model's generalization to different data. To address this issue, an alternative approach represented in Fig. \ref{code:preprocessing_leakage}b involves computing the function parameters only on the training data $D^T$, denoted as $Q^T = \upsilon(D^T)$, after the dataset split. The preprocessing function is then applied to both $D^T$ and $D^E$ using the same parameter set $Q^T$. Consequently, the training stage is influenced only by information in the training data, leading to a more reliable evaluation of the model's generalization on novel instances, as it is less prone to overfitting to specific characteristics of the entire dataset. }

\begin{figure}
    \centering
    \input{_IMG_CODE_PREPROCESSING_LEAK}
    \label{code:preprocessing_leakage}
\end{figure}

This step helps prevent potential over-fitting and overoptimistic performance assessments during evaluation. For instance, the authors of \citep{kapoor2023leakage} replicated the experimental setups  of some literature studies influenced by imputation leakage, reporting results different from the ones made by the original authors. The authors in \cite{bouke2023empirical} investigated how incorrect standardization procedures can result in leakage across several ML methods applied on established datasets used for bench-marking intrusion detection systems. \newtexttwo{The authors of \cite{whalen2022navigating} named this form of data leakage in genomics as \textit{leaky preprocessing}, highlighting how it frequently occurs in the literature. They emphasize that many common transformations, such as standardization, PCA, and other common preprocessing methods, can unintentionally introduce information from the test set into the training process, thereby compromising the validity of model evaluation. Similar conclusions was reported in \citep{rocke2009papers} for feature selection on microarray data.} 
\newtext{It is important to emphasize that a correct split should be made not only between training data and final evaluation data (commonly named test set), but between training set and every out-of-the-training data involved in the evaluation.} Indeed, considering the scenario described in Sec. \ref{sec:preprop:particular}, while it's a common practice to reserve a portion of the training data as validation data (i.e. $D^{T}= D^{T'} \cup D^{V} \text{ and } D^{T'} \cap D^{V} = \emptyset$, where $D^{T'}$ is the dataset actually used to train the model) for early-stopping, this strategy, while seemingly safe, might not completely prevent \textit{before-split leakage} cases due to operations, such as preprocessing parameter computations, performed on the entire $D^{T}$ before the partition of $D^{T}$ in $D^{T'} \text{ and } D^{V}$ is made.  Therefore, the risk is not only biasing the training stage but also potentially returning a model not properly trained. 
This problem can easily arise among practitioners due to the current implementation of the learning procedures in existing ML packages. These packages often provide the option to specify how many training data instances should be used as the validation set, without taking into account whether the training data has been influenced by any prior transformations\footnote{see for example the model \textit{fit} function provided by the $Keras$ package \url{https://keras.io/api/models/model_training_apis/} (last seen on 12/01/2024).}. 
However, this kind of leakage can be easily avoided considering also the validation set $D^{V}$ (such as every evaluation set) as part of the splitting strategy, and considering only $D^{T'}$ to compute preprocessing parameters. This scenario is summarized in  Fig. \ref{code:validation_leak}. \newtexttwo{In particular, Let's $train^{VA}$ a training function adopting a model $M$ to train, a training set $D^{T'}$ and a validation set $D^V$ to monitor the model performance during the training. If $D^V$ is taken as part of the $D^T$ set as in Fig. \ref{code:validation_leak}a, a leakage between validation and training data can arise if data management or feature engineering are made on the whole $D^T$. Consider a preprocessing function (see for example Fig. \ref{code:preprocessing_leakage}) or dataset synthesis functions (see for example Fig. \ref{code:synthesis_leakage}). Differently from the scenario depicted in Fig. \ref{code:preprocessing_leakage}b, where the risk of leakage is mitigated, in the current scenario there is an additional validation set adopted during the training stage, usually obtained from $D^T$. Here, the training set $D^{T'}$ may be preprocessed using information from the $D^V$, potentially biasing the training stage. The model training might be prematurely stopped, satisfied with its performance on $D^V$ but failing to generalize well to $D^E$ (see Fig. \ref{fig:validation_leak}). This risk emerges because $D^V$ may not accurately represent out-of-the-training data due  to, for example, the influence of preprocessing functions conducted with parameters computed on the entire set $D^{T}$. Therefore, the training stop criteria based on $D^V$ may not correspond to the real model generalization capabilities. A possible solution represented in Fig. \ref{code:validation_leak}b is to split the data between actual training data $D^{T'}$ and $D^{V}$ \textit{before} the preprocessing (or other functions exploiting knowledge on the whole input data) and compute any preprocessing parameters (if applicable) only on $D^{T'}$. }
\begin{figure}
\input{_IMG_CODE_VALIDATION_LEAK}
\label{code:validation_leak}
\end{figure}

Finally, depending on the chosen ML paradigm, i.e. inductive or transductive, the experimental assessment may or may not be impacted by this type of leakage.
Similar consideration can be made for TL cases. 
Focusing on the preprocessing step, model evaluation should be made exclusively on data originating from the target domain not employed to compute preprocessing parameters or to make feature engineering operations. To ensure accurate evaluation, a portion of target data needs to be reserved and untouched during the training phase, for evaluation purposes. This is also to be considered in supervised DA cases. Indeed, supervised DA strategy assumes the availability of labeled target data specific for the new task. Take, for instance, a scenario involving a pre-trained model. This model can be seen as a new model starting from a more advantageous initial state given by the pre-training stage (for example, having better initial weight values in case of ANNs). Consequently, assuming that the new dataset is totally different from the dataset used to pre-train the model, the risks of leakage before the data splitting stage remain similar to those in a standard ML approaches. For instance, this could involve computing normalization parameters using information from the entire new dataset instead of restricting it solely to the target training data.

\subsection{\textbf{Split-related leakage}}
\label{sec:dataleaktypes:splitrelated}
In this section we will describe cases of leakage which can happen during the split between training and out-of-the-training data has been made (see Fig. \ref{fig:leakage_type}). Indeed, despite the fact that a separation between datasets is made, leak of information can be raised during the split strategy. 
In the following, we refer to a split strategy as a function $h: \mathbb{D} \to \mathbb{D} \times \mathbb{D} \times \dots \times \mathbb{D}$ that, given a dataset $D \in \mathbb{D}$, returns a collection of datasets $\{D_j\}_{j=1}^J$ derived from $D$. Usually, $D^T \leftarrow D_a$ and $D^E \leftarrow D_b$, with $a,b\in {1,2,\dots,J}$.
  

Scenarios involved in Split-related leakage can be categorized into two main types: 
\newtext{\begin{enumerate}[i.]
    \item \textit{Similarity leakage}: when the training and out-of-the-training contain a part or all of the same data;
    \item \textit{Distribution leakage}: when the split doesn't care about the different distributions of the data, introducing knowledge in the training stage not representative of the real task. 
\end{enumerate}}
Differently from similar type of leakage such as data collecting leakage, distribution leakage is easily avoidable adopting a proper split strategy. Indeed, as highlighted in Sec. \ref{sec:dataleaktypes:inner}, in this work we consider as ``data collecting leakage'' the cases when it is not possible to avoid leakage through improved split strategies. This occurs because the information is inherently contaminated at the source, thereby compromising the integrity of the data instance features. 
 
\subsubsection{Similarity leakage} 
 \label{sec:dataleaktypes:similarity}
This type of leakage can happen when there is some kind of similarities or connection between data points in different sets. 
\newtext{We can distinguish between:
\begin{enumerate}[i.]
\item \textit{Sets' intersection leakage}: if the training and out-of-the-training contain a part or all of the same data;
\item \textit{Overlap leakage}: when data features of different data instances overlap between them.
\end{enumerate}
}

\paragraph{Sets' intersection leakage}
\label{sec:dataleaktypes:setsintersect}
In this case, the leakage happens if the split strategy $h$ assign the same data to both $D^T$ and $D^E$ in a way not allowed by the task. 
This can be happens in case of particular sampling strategy for $D^T$ and $D^E$. 

\newtext{Although apparently easy to avoid, a specific instance of this type of leakage can easily occur when several evaluation set are used, for example in cases that rely on a validation dataset $D^{V}$ to monitor model performance during training, similarly to the case described in Sec. \ref{sec:preprop:particular}.
Two possible scenarios are outlined below, the former describe a very simple case where the split strategy uses an improper sampling procedure, the latter describe how the adoption of an early stopping criteria and a set's intersection split strategy can lead to a leakage condition.} 
\begin{figure}
\centering
\begin{tikzpicture}

\definecolor{darkgray176}{RGB}{176,176,176}
\definecolor{lightgray204}{RGB}{204,204,204}
\definecolor{orange}{RGB}{255,165,0}

\begin{axis}[
legend cell align={left},
legend style={
  fill opacity=0.8,
  draw opacity=1,
  text opacity=1,
  at={(0.03,0.03)},
  anchor=south west,
  draw=lightgray204
},
scaled x ticks=manual:{}{\pgfmathparse{#1}},
scaled y ticks=manual:{}{\pgfmathparse{#1}},
tick align=outside,
title={Illustration of Sets' Intersection Leakage},
x grid style={darkgray176},
xlabel={\(\displaystyle x_1\)},
xmajorgrids,
xmajorticks=false,
xmin=-2.91128808863977, xmax=2.63526062550698,
xtick style={color=black},
xticklabels={},
y grid style={darkgray176},
ylabel={\(\displaystyle x_2\)},
ymajorgrids,
ymajorticks=false,
ymin=-3.02405856911289, ymax=2.50818930998445,
ytick style={color=black},
yticklabels={}
]
\addplot [draw=blue, fill=blue, mark=*, only marks]
table{%
x  y
-0.672460447775951 -0.359553161540541
-0.813146282044454 -1.72628260233168
0.177426142253753 -0.401780936208262
-1.63019834696604 0.462782255525774
-0.907298364383242 0.051945395796139
0.729090562177537 0.128982910757411
1.1394006845433 -1.23482582035365
0.402341641177549 -0.684810090940313
-0.870797149181882 -0.578849664764415
-0.311552532127373 0.0561653422297454
-1.16514984078336 0.900826486954187
0.46566243973046 -1.53624368627722
1.4882521937956 1.89588917603058
1.17877957115965 -0.179924835812351
-1.07075262151054 1.05445172693114
-0.40317694697318 1.22244507038243
0.20827497807686 0.976639036483713
0.356366397174402 0.706573168191948
0.0105000207208205 1.78587049390584
0.12691209270362 0.401989363444702
1.88315069705625 -1.34775906114245
-1.27048499848573 0.969396708158011
-1.17312340511416 1.94362118564929
-0.413618980759747 -0.747454811440758
1.92294202648038 1.48051479143442
1.86755896042657 0.906044658275385
-0.861225685054703 1.91006495309903
-0.26800337095138 0.802456395796395
0.947251967773748 -0.155010093090834
0.61407937034608 0.922206671566527
0.376425531155629 -1.09940079058419
0.298238174206056 1.32638589668703
-0.694567859731365 -0.149634540327671
-0.435153551721637 1.84926372847934
0.672294757012435 0.40746183624111
-0.769916074445316 0.539249191291817
-0.674332660657376 0.0318305582743512
-0.635846078378881 0.6764332949465
0.576590816614941 -0.208298755577995
0.396006712661645 -1.09306150873051
-1.49125759270561 0.439391701264537
0.166673495372529 0.635031436892106
2.38314477486394 0.944479486990414
-0.912822225444159 1.11701628809585
-1.31590741051152 -0.461584604814709
-0.0682416053246312 1.71334272164937
-0.74475482204844 -0.826438538659014
-0.0984525244254323 -0.663478286362107
1.12663592210651 -1.07993150836342
-1.1474686524111 -0.437820044744434
-0.498032450692305 1.92953205381699
0.949420806925761 0.0875512413851909
-1.22543551883017 0.844362976401547
-1.00021534738956 -1.54477109677761
1.1880297923523 0.31694261192485
0.920858823780819 0.318727652943021
0.856830611902691 -0.651025593300147
-1.03424284178446 0.681594518281627
-0.803409664173841 -0.689549777750201
-0.455532503517343 0.0174791590250567
-0.353993911253484 -1.37495129341802
-0.643618402832891 -2.22340315222443
0.625231451027187 -1.60205765560675
-1.10438333942845 0.0521650792609744
-0.739562996391313 1.54301459540674
-1.29285690972345 0.267050869349183
-0.0392828182274956 -1.1680934977412
0.523276660531754 -0.171546331222248
0.771790551213667 0.823504153963731
2.16323594928069 1.33652794943639
1.76405234596766 0.400157208367223
0.978737984105739 2.24089319920146
1.86755799014997 -0.977277879876411
0.950088417525589 -0.151357208297698
-0.103218851793558 0.410598501938372
0.144043571160878 1.45427350696298
0.761037725146993 0.121675016492828
0.443863232745426 0.333674327374267
1.49407907315761 -0.205158263765801
0.313067701650901 -0.854095739301725
-2.55298981583408 0.653618595440361
0.864436198859506 -0.742165020406442
2.26975462398761 -1.45436567459876
0.0457585173014461 -0.187183850025834
1.53277921435846 1.46935876990029
0.154947425696916 0.378162519602174
-0.887785747630113 -1.98079646822393
-0.347912149326153 0.15634896910398
1.23029068072772 1.20237984878441
-0.387326817407952 -0.302302750575336
-1.04855296506709 -1.42001793717898
-1.70627019062501 1.95077539523179
-0.509652181751653 -0.438074301611186
-1.25279536004993 0.77749035583191
-1.61389784755795 -0.212740280213969
-0.895466561193676 0.386902497859262
-0.510805137568873 -1.18063218412241
-0.0281822283386549 0.428331870530418
0.0665172223831679 0.302471897739781
-0.634322093680964 -0.362741165987138
};
\addlegendentry{$D^T$}
\addplot [draw=orange, fill=orange, mark=square*, only marks]
table{%
x  y
-0.369181837942444 -0.239379177575926
1.09965959588711 0.655263730722598
0.640131526097592 -1.61695604431083
-0.0243261243989356 -0.738030909205689
0.279924599043238 -0.0981503896429579
0.910178908092592 0.317218215191302
0.786327962108976 -0.466419096735943
-0.94444625591825 -0.410049693202548
-0.0170204138614406 0.379151735555082
2.25930895069085 -0.0422571516606427
-0.955945000492777 -0.345981775699386
-0.463595974646094 0.481481473773462
-1.54079701444462 0.0632619942003317
0.156506537965376 0.232181036200276
-0.597316068965363 -0.237921729736007
-1.42406090898253 -0.493319883362194
-0.542861476016718 0.416050046261425
-1.15618243182191 0.781198101709993
1.49448454449137 -2.06998502501353
0.426258730778101 0.676908035030246
-0.637437025552229 -0.397271814328798
-0.132880577586956 -0.297790879401728
-0.309012969047122 -1.67600380632998
1.15233156478312 1.07961859203682
-0.813364259204203 -1.46642432780251
0.521064876452759 -0.575787969813066
0.14195316332078 -0.319328417145095
0.691538751070187 0.694749143656006
-0.725597378463584 -1.38336395539506
-1.58293839733508 0.610379379107205
-1.18885925778403 -0.506816354298688
-0.596314038450508 -0.0525672962695463
-1.93627980584651 0.188778596793829
0.523891023834206 0.0884220870446614
-0.310886171698472 0.0974001662687834
0.39904634564013 -2.77259275642665
1.95591230825069 0.390093322687926
-0.65240858238702 -0.390953375187601
0.493741777349188 -0.116103939034367
-2.03068446778149 2.06449286135932
-0.110540657232473 1.0201727117158
-0.692049847784391 1.5363770542458
0.28634368889228 0.608843834475451
-1.04525336614695 1.2111452896827
0.689818164534788 1.301846229565
-0.628087559641579 -0.481027118460788
2.30391669768394 -1.06001582272155
-0.135949700678321 1.1368913626027
0.0977249677148556 0.582953679753294
-0.399449029262875 0.370055887847519
-1.30652685173532 1.65813067961819
-0.11816404512857 -0.68017820399685
0.666383082031914 -0.460719787388553
-1.33425847140275 -1.34671750579756
0.693773152690133 -0.159573438146267
-0.133701559668439 1.07774380597626
-1.12682580875674 -0.730677752864825
-0.384879809181275 0.094351589317074
-0.0421714512905789 -0.286887192389908
-0.0616264020956474 -0.107305276291175
-0.719604388551793 -0.812992988554077
0.27451635772394 -0.890915082995528
-1.15735525919085 -0.312292251125693
-0.157667016163816 2.25672349729821
-0.704700275856234 0.943260724969495
0.747188334204632 -1.18894495520374
0.7732529774026 -1.18388064019332
-2.65917223799674 0.606319524359381
-1.75589058343772 0.450934461805915
-0.684010897737217 1.65955079618987
1.76405234596766 0.400157208367223
0.978737984105739 2.24089319920146
1.86755799014997 -0.977277879876411
0.950088417525589 -0.151357208297698
-0.103218851793558 0.410598501938372
0.144043571160878 1.45427350696298
0.761037725146993 0.121675016492828
0.443863232745426 0.333674327374267
1.49407907315761 -0.205158263765801
0.313067701650901 -0.854095739301725
-2.55298981583408 0.653618595440361
0.864436198859506 -0.742165020406442
2.26975462398761 -1.45436567459876
0.0457585173014461 -0.187183850025834
1.53277921435846 1.46935876990029
0.154947425696916 0.378162519602174
-0.887785747630113 -1.98079646822393
-0.347912149326153 0.15634896910398
1.23029068072772 1.20237984878441
-0.387326817407952 -0.302302750575336
-1.04855296506709 -1.42001793717898
-1.70627019062501 1.95077539523179
-0.509652181751653 -0.438074301611186
-1.25279536004993 0.77749035583191
-1.61389784755795 -0.212740280213969
-0.895466561193676 0.386902497859262
-0.510805137568873 -1.18063218412241
-0.0281822283386549 0.428331870530418
0.0665172223831679 0.302471897739781
-0.634322093680964 -0.362741165987138
};
\addlegendentry{$D^E$}
\addplot [draw=black, fill=red, mark=*, only marks]
table{%
x  y
1.76405234596766 0.400157208367223
0.978737984105739 2.24089319920146
1.86755799014997 -0.977277879876411
0.950088417525589 -0.151357208297698
-0.103218851793558 0.410598501938372
0.144043571160878 1.45427350696298
0.761037725146993 0.121675016492828
0.443863232745426 0.333674327374267
1.49407907315761 -0.205158263765801
0.313067701650901 -0.854095739301725
-2.55298981583408 0.653618595440361
0.864436198859506 -0.742165020406442
2.26975462398761 -1.45436567459876
0.0457585173014461 -0.187183850025834
1.53277921435846 1.46935876990029
0.154947425696916 0.378162519602174
-0.887785747630113 -1.98079646822393
-0.347912149326153 0.15634896910398
1.23029068072772 1.20237984878441
-0.387326817407952 -0.302302750575336
-1.04855296506709 -1.42001793717898
-1.70627019062501 1.95077539523179
-0.509652181751653 -0.438074301611186
-1.25279536004993 0.77749035583191
-1.61389784755795 -0.212740280213969
-0.895466561193676 0.386902497859262
-0.510805137568873 -1.18063218412241
-0.0281822283386549 0.428331870530418
0.0665172223831679 0.302471897739781
-0.634322093680964 -0.362741165987138
};
\addlegendentry{Shared}
\end{axis}

\end{tikzpicture}
\caption{\newtexttwo{Sets' intersection leakage: some data instance are included in both $D^T$ and $D^E$ by the adopted split strategy.}}
\label{fig:set_intersection}

\end{figure}

\casestudybox{A Sets' intersection leakage due to sampling procedures}{Suppose we have a dataset for a classification problem. To split the dataset into training and evaluation sets, the split strategy $h$ adopts a sampling procedure with reintroduction to generate the output datasets. Therefore, this random sampling strategy leads to some data instance being included in both $D^T$ and $D^E$, as shown in Fig. \ref{fig:set_intersection}. 
Such overlap constitutes leakage if the task does not inherently allow identical instances to appear across training and test sets.}

\casestudybox{A sets' intersection leakage impacting in early stopping criteria}{Suppose we have a dataset splitted into $D^{T}$ and $D^{E}$. We want to training a neural network model on $D^{T}$ using Gradient Descent, where the model iteratively updates its parameters in multiple steps. The training process is monitored using a validation set $D^{V}$ to implement early stopping, a mechanism that halts training when performance on the validation set no longer improves, preventing overfitting. Specifically, at each iteration of Gradient Descent, the model's performance on $D^{V}$ is evaluated, and if there is no significant improvement, the training stops early. The model's parameters are iteratively updated only using the training set $D^{T}$, and the validation set is only used to determine when to stop training. 
Since the validation set does not directly affect the model parameters, one could set $D^{V}=D^{E}$. The final model will be evaluated on $D^{E}$. 
But, since the validation set includes data from the test set, the model's stopping point is influenced by data it should not have access to during training. When the model performs well on $D^{V}=D^{E}$, it signals the end of training,  proposing a model configuration fitted for $D^{E}$, generating inflated evaluation.}

\paragraph{Data overlap leakage}
\label{sec:dataleaktypes:overlap}
This case can occur during the design phase of the data instances, specifically when determining which features will constitute the data instances, i.e. 

\begin{definition} \newtexttwo{Data overlap leakage can occur if a split strategy $h$ applied on $D$ to obtain the $D^T$ and $D^E$ results in data segments shared between $D^T$ and $D^E$, i.e. $\exists (\vec{x}^{(i)},y^{(i)}) \in D^T, \exists (\vec{x}^{(l)},y^{(l)}) \in D^E$ s.t. $x_t^{(l)}\leftarrow x^{(i)}_u \lor  x^{(i)}_u \leftarrow x_t^{(l)}$ for some $t,u \in \{1,2,\dots,d\}$.}  
\end{definition}

Unlike data collection leakage, which can occur during the data acquisition process, this kind of leakage can originate during the data preparation phase, after that the data has already been acquired, when it is being organized into the final dataset. \newtext{Then, the data instance design, combined with the chosen split strategy, can contribute to leakage.}

An example is provided in the following scenario.

\casestudybox{A data overlap leakage due to shared features}{Consider a dataset of EEG signals acquired from one or several subjects. In order to capture the temporal dynamics of brain signals, the continuous EEG data is often segmented into smaller time windows (\textit{epochs}) representing a short window of the EEG signal, and overlapping windows are often used. In this case, the EEG data is segmented into overlapping epochs with a 50 \% overlap, meaning that consecutive epochs share half of the same data points.
Suppose that EEG epochs are used to train a ML model for classifying different brain emotions. 
When splitting the data into training and out-of-training sets, random sampling is applied to the epochs without considering the overlap between them. This means that some epochs in $D^T$ and the out-of-training sets can share a significant portion of the same EEG signal data due to the overlap. Therefore, the model is being trained on epochs that are very similar to those in $D^E$, despite the splitting. Since the model has already seen much of the data from the overlapping segments during the training, the evaluation results does not reflect the model's true ability to generalize to completely unseen data.}
\newtexttwo{Park and Marcotte \citep{park2012flaws} showed that in pair-input settings, where models operate on pairs of objects rather than on individual instances, standard split strategies such as randomly splitting pairs into training and evaluation sets can lead to data leakage due to overlap at the level of individual components.}

\subsubsection{Distribution leakage} 
\label{sec:dataleaktypes:distribution}
This type of leakage occurs when different distributions of the data introduce some kind of knowledge during the training which can affect the final performance evaluation. 
\begin{definition}
\newtexttwo{Let $T$ be a task in which the input data is naturally drawn from multiple underlying distributions $\mathcal{P}_1, \mathcal{P}_2, \dots, \mathcal{P}_k$. Let the available dataset be $D = \bigcup_{i=1}^k D^{i}$, with each $D^i \sim \mathcal{P}_i$. Let $O$ be a data splitting operation such that $O(D) = \{D^T, D^E\}$ and let $P_T$, $P_E$ denote the empirical distributions induced by $D^T$ and $D^E$, respectively.
We say that the operation $O$ introduces a split-induced distribution leakage if the resulting $D^T$ and $D^E$ are unrepresentative of the distributional heterogeneity $\mathcal{P}_1, \mathcal{P}_2, \dots, \mathcal{P}_k$ assumed by task $T$.}
\end{definition}


If the split between training and evaluation sets is made without considering the data sources and the task, it is possible that information contained in the training data can pollute in an improper way the evaluation dataset instances coming from the same sources. 
A possible instance of this type of leakage can arise mixing together the data coming from different sources both in training and evaluation datasets, not taking into account the different sources from which the data comes. Thus, the ML model could have effectively learned to perform well on data instances coming from familiar sources, struggling when faced with new or different sources not encountered during training. For example, the model might exploit biases specific to certain data used during the training, such as noise or artifacts. Another case of distribution leakage can arise when data instances represent features with a sequential relationship. For example, when data collected at different time points exists in the training and out-of-the-training sets, and data acquire at a given time helps to infer the data acquired in other times. 
In the following we report possible scenarios from the literature of these types of leakage.

\casestudybox{A Distribution leakage due to shared distribution between data instances}{In \cite{backstrom2018efficient}, a study for automatic feature extraction and classification of Alzheimer's Disease using 3D MRI brain scans as input is proposed. The classification system is tested on an ``Alzheimer's Disease Neuroimaging Initiative'' dataset\footnote{\url{http://adni.loni.usc.edu}}, which contains approximately 1200 MRI scans from 340 patients. In one of the experimental setups, the dataset is randomly partitioned into training and test sets without taking into account the patients from whom the MRI scans were acquired. However, each participant can be considered a distinct source, with their data following a unique distribution. If evaluation data comprises instances from the same distributions as the training data, the model might exploit artifacts or biases specific to certain sources, resulting in biased evaluations and misleading performance metrics.} 
The same study repeats the experiments adopting also a subject-separated MRI data partitioning strategy, with a drop in accuracy of about 8 \%.
However, Depending on the task, overlooking on the adoption of data from different participants in the evaluation set(s) can be a case of data leakage or not. Intuitively, using data coming from different distributions during the training can lead to a model with greater generalization capability, since the model can learn from data instances very different between them. Conversely, if evaluation data comprises instances from the same distributions as the training data, the model might exploit artifacts or biases specific to certain sources, resulting in biased evaluations and misleading performance metrics. This topic is discussed in \citep{kamrud2021effects} about studies involving EEG signals. In particular, the evaluation made on subject-separated partitioning exhibiting lower effectiveness respect to naive random partitioning. In real-world scenarios, where new subject scans are usually unseen by the ML system, partitioning schemes not considering subjects as distinct entities may be prone to data leakage.

\casestudybox{A distribution leakage due to temporal dependence between data instances}{Consider a  dataset of daily stock prices for the last five years. Consider the task of predict stock prices based on historical data adopting a ML model.
The dataset $D$ is splitted into training and evaluation sets $D^T$ and $D^E$ using a random sampling strategy, where data from different time periods is mixed across both sets. As a result, the model might be trained on stock prices from July 2020 randomly placed in $D^T$ and evaluated on prices from January 2020 (i.e., previous data) randomly placed in $D^E$. In this case, stock prices from July (used in training) may contain important information about market trends that started developing in January, unintentionally helping the model to correctly predicting June's prices during model evaluation.
More in general, information obtained at one time can inadvertently assist in inferring or predicting data acquired at other time points, such as evaluating the model on data that precedes the data used for training.}

\subsubsection{Split-related leakage in Transductive Machine Learning and Transfer Learning}
\newtexttwo{In transductive ML, the model is trained with the explicit goal of making predictions on a fixed, known set of input instances, rather than generalizing to unseen future data. This setting allows the model to access the entire set of inputs—both labeled and unlabeled—during training. However, even in this context, certain forms of data leakage may still arise. These do not stem from a violation of the inductive ML assumptions, but rather from inconsistencies between the task definition and the structure of the data used for training and evaluation. For instance, if the task assumes that instances are independent, but the training and test sets contain highly similar or overlapping examples due to the split strategy such as duplicated segments, repeated measurements, or chunks from the same source, then the model may exploits these similarities, rather than learning meaningful associations. This type of leakage, while not violating the transductive setting per se, can still invalidate the intended evaluation and compromise the reliability of the model’s predictions. In the following a possible scenario of this case. In the following part of this section, a transductive ML split-related leakage scenario is described, followed by two split-related leakage respectively in the DA context and in the DG context.}

\label{sec:dataleaktypes:splitrelated:trans}
\casestudybox{A split-related leakage in a Transductive ML due to learning input-specific patterns}{\newtexttwo{A transductive model is trained to classify short text excerpts (chunks) into categories such as “Legal”, “Technical”, or “Marketing”. The entire set of unlabeled chunks is known in advance, and a subset of them is manually labeled for training. The model is expected to predict the labels of the remaining ones.
Chunks originating from the same document appear both in the labeled and unlabeled sets. Although all inputs are accessible (as permitted in transductive learning), the model learns to associate document-specific patterns rather than category-level patterns.}}

\casestudybox{A Distribution leakage in DA due to learning source-specific patterns}{Suppose we have a classification problem consisting in predicting emotions from a dataset of EEG acquisition. The dataset is composed of EEG recordings from multiple subjects. Since the signals are significant non-stationary due to individual differences and varying temporal conditions, we employ a DA strategy by using data $D_{target}$ from a target domain (EEG recordings from the subjects we want to generalize to) and data $D_{source}$ source domain (EEG recordings from a different set of subjects).
During the training process, since DA methods assume that data instances $\vec{x}$ from target domain dataset $D_{source}$ are available during the training, the model extensively use them (e.g., EEG data from the evaluation subject) to adjust the model parameters. However, in the case that a portion of the target domain data instances has not been reserved for the final evaluation, we have data-leakage. 
}
More in general, by allowing the model to be trained from the whole target domain data, it becomes tailored to the specific data collected. This means the model could not generalize well to new data, also belonging the same target domain. This scenario is similar to a semi-supervised standard ML scenario, but with different data distributions and/or tasks involved.

\casestudybox{A Split-related leakage in DG due to learning source-specific patterns}{Suppose we are working on a classification task where the goal is to predict emotions using fMRI data acquired from $S$ subjects considered as different domains, i.e. $D=\bigcup\limits_{s=1}^{S}D_s$. The goal is to build a model that generalizes to new, unseen subjects $s' \in S$ (the target domain) to assess its inter-subject generalization capability.
The evaluation set is composed of all the data belonging to a specific subject, i.e. $D^E = D_{s'}$  for some $s' \in \{1,2,\dots, S\}$, and the training set of the remaining data $D^T=D\setminus D_{s'}$.
However, distribution leakage can occur if the target subject is closely related to one of the subjects in the source domain. For example, if data recorded by siblings or twins are present in the dataset. Because brain structure and function are known to be heritable \cite{rosenblatt2024data}, having one family member in the training set and another in the test set could allow the model to exploit familial similarities rather than learning generalizable brain-behavior relationships. 
This does not reflect true generalization, since the model overfits to the subject-specific characteristics in the recorded fMRI failing to generalize to new subjects.}

\subsubsection{Split-related leakage: Discussion}
\label{sec:dataleaktypes:splitrelated:general_disc}
To give an idea the impact of set's Intersection Leakage, we designed a binary synthetic classification task where a portion of the test set is inadvertently included in the training set—simulating a common data handling mistake where the train/test separation is violated.

We generated synthetic datasets using a binary classification benchmark. For each experiment, we performed stratified 5-fold cross-validation and progressively increased the percentage of test samples that were duplicated and inserted into the training set, ranging from 0 \% (clean split) to 100 \% (total contamination). A simple two layer DNN with 100 and 50 neurons respectively was used for evaluation.

The experiment was repeated 10 times to account for variability, and we report the average classification accuracy and standard deviation at each leakage level.

The results, shown in Fig. \ref{fig:exp:set_intersection_leakage}, reveal that even low contamination of the training set with test data can lead to artificially inflated performance estimates, highlighting the need for strict separation protocols when partitioning datasets.

\begin{figure}
\centering
\begin{tikzpicture}

\definecolor{darkgray176}{RGB}{176,176,176}

\begin{axis}[
tick align=outside,
tick pos=left,
title={Effect of Set's Intersection Leakage on Accuracy},
x grid style={darkgray176},
xlabel={Percentage of test set copied into training set},
xmajorgrids,
xmin=-0.5, xmax=10.5,
xtick style={color=black},
xtick={0,1,2,3,4,5,6,7,8,9,10},
xticklabels={\tiny 0\%,\tiny  10\%,\tiny 20\%,\tiny 30\%,\tiny 40\%,\tiny 50\%,\tiny 60\%,\tiny 70\%,\tiny 80\%,\tiny 90\%,\tiny 100\%},
y grid style={darkgray176},
ylabel={Accuracy},
ymajorgrids,
ymin=0, ymax=1.05,
ytick style={color=black}
]
\path [draw=blue, semithick]
(axis cs:0,0.893554925408249)
--(axis cs:0,0.946845074591752);

\addplot [semithick, blue, mark=-, mark size=5, mark options={solid}, only marks]
table {%
0 0.893554925408249
};
\addplot [semithick, blue, mark=-, mark size=5, mark options={solid}, only marks]
table {%
0 0.946845074591752
};
\path [draw=red, semithick]
(axis cs:1,0.896678790587439)
--(axis cs:1,0.950921209412561);

\path [draw=red, semithick]
(axis cs:2,0.903143370215207)
--(axis cs:2,0.954056629784793);

\path [draw=red, semithick]
(axis cs:3,0.905746977142758)
--(axis cs:3,0.957853022857242);

\path [draw=red, semithick]
(axis cs:4,0.910297940154363)
--(axis cs:4,0.963702059845637);

\path [draw=red, semithick]
(axis cs:5,0.918828816607938)
--(axis cs:5,0.968771183392062);

\path [draw=red, semithick]
(axis cs:6,0.92669242548427)
--(axis cs:6,0.97290757451573);

\path [draw=red, semithick]
(axis cs:7,0.934594898623062)
--(axis cs:7,0.973805101376937);

\path [draw=red, semithick]
(axis cs:8,0.942665675368166)
--(axis cs:8,0.978534324631834);

\path [draw=red, semithick]
(axis cs:9,0.947805618864372)
--(axis cs:9,0.983394381135628);

\path [draw=red, semithick]
(axis cs:10,0.9568)
--(axis cs:10,0.9876);

\addplot [semithick, red, mark=-, mark size=5, mark options={solid}, only marks]
table {%
1 0.896678790587439
2 0.903143370215207
3 0.905746977142758
4 0.910297940154363
5 0.918828816607938
6 0.92669242548427
7 0.934594898623062
8 0.942665675368166
9 0.947805618864372
10 0.9568
};
\addplot [semithick, red, mark=-, mark size=5, mark options={solid}, only marks]
table {%
1 0.950921209412561
2 0.954056629784793
3 0.957853022857242
4 0.963702059845637
5 0.968771183392062
6 0.97290757451573
7 0.973805101376937
8 0.978534324631834
9 0.983394381135628
10 0.9876
};
\addplot [semithick, blue, mark=o, mark size=3, mark options={solid,fill opacity=0}, only marks]
table {%
0 0.9202
};
\addplot [semithick, red, mark=x, mark size=3, mark options={solid}, only marks]
table {%
1 0.9238
2 0.9286
3 0.9318
4 0.937
5 0.9438
6 0.9498
7 0.9542
8 0.9606
9 0.9656
10 0.9722
};
\end{axis}

\end{tikzpicture}

\caption{\newtexttwo{The x-axis indicates the percentage of test samples erroneously duplicated into the training set, simulating a common data handling error where training and test sets are not properly isolated. The y-axis shows the classification accuracy. A clean split (blue) serves as the baseline. As the overlap increases (red), performance improves artificially due to the model encountering test instances during training.}}
\label{fig:exp:set_intersection_leakage}
\end{figure}

Sets' intersection leakage can be easily managed avoiding data points intersections during the dataset splits, as represented in Fig. \ref{code:intersection_leak}. \newtexttwo{In particular, in Fig. \ref{code:intersection_leak}a the data are simply divided between training $D^T$ and evaluation $D^E$ sets without taking care to any intersection between set (for example due to replicates). A corrected version represented in \ref{code:intersection_leak}b can be obtained constraining the split strategy to build $D^T$ and $D^E$ without intersection.}
\begin{figure}
    \centering    \input{_IMG_CODE_DATASET_INTERSECTION_LEAK}
    \label{code:intersection_leak}
\end{figure} 
Intersection between training and test data is not always considered wrong in literature. A scenario can be found in \citep{lee2018validity}, focusing on classifying the pen model or brand of an ink entry. The study conducts a performance comparison between two potential configurations of training and test sets. In one configuration, ink strokes originating from a specific pen are exclusively included in either the training or test set. In the other configuration, ink strokes data from a particular pen can be included in either the training or test sets without any constraints, resulting in replicates spread between the training and the test sets. Nevertheless, the authors deem the performance obtained with the second configuration acceptable. This acceptance stems from the task at hand, which is to aid forensic ink analysts in identifying the pen brand or model of a questioned ink stroke, rather than pinpointing the exact pen responsible. Consequently, considering that a high-quality evaluation set should mirror the specific data structure characteristics of the training set to closely resemble future samples, the presence of duplicates in both the training and test sets is considered acceptable. This scenario highlights the importance in defining the task to understand if there is a leakage problem or not. \newtext{Furthermore, it is important to consider if other evaluation sets different from the final evaluation set (test set) are involved, as described in the scenario in Sec. \ref{sec:dataleaktypes:setsintersect}. More in general, adopting (part of) $D^{E}$ as (part of) a validation set $D^{V}$ (i.e. $D^{V} \cap D^{E} \neq \emptyset$) during the training process for early stopping doesn't directly impact the model because the ML model parameters learn solely from $D^{T}$, while $D^{V}$ is only used to select the model parameters corresponding to the best generalization capability of the model. $D^{V}$ is thus usually used to implement a stopping criterion that allows a good trade-off between generalization capability and computational cost. However, $D^{V} \cap D^{E} \neq \emptyset$ can lead to stop the training process when the learned model reports good performance on data belonging to the test set, again resulting in overoptimistic generalization performance. Therefore, the optimal scenario corresponds to completely disjoint validation and test sets.}

Finally, the context-specific nature of split-related leakage emphasizes the necessity of aligning evaluation goals with the intended use of the final model and dataset construction. In particular, four factors can lead to a split-related leakage: The task, the dataset construction, the learning procedure, and the split strategy. These factors are interconnected between them and mismanaging any of these factors can result in a situation where one or more leakage scenarios occur. For example, scenarios as data collecting leakage (see Sec. \ref{sec:dataleaktypes:inner}) can propagate their effects after the data split (see for example Sec. \ref{sec:dataleaktypes:overlap}). However, while split-related not due to data-induced leakage can be usually solved with proper split strategy, Data-induced leakage could be harder to solve due to the pitfalls in the acquired data. 

The connection between split strategy and task involved is particularly evident in distribution leakage: The task can determine if the chosen split strategy can lead to a leakage condition or not. An example is provided in \citep{kamrud2021effects} about ML task involving EEG data. Specifically, if a ML model working on EEG is designed solely for classification within the same population it is trained upon, then testing the model with unseen individuals from the same population may be sufficient. In this case, the emphasis is on capturing population-specific characteristics. Conversely, when the ML model is intended for broader applications and generalization to unseen individuals (representing the general population) the data should be partitioned to ensure that individuals used for training are distinct from those used for validation and testing. \newtexttwo{\citep{brookshire2024data} discussed about this type of leakage, highlighting that to compensate for the limited size of  EEG datasets, many studies artificially increase the number of training samples by splitting a single recording into multiple segments. However, when these segments from the same subject appear in both the training and test sets, it results in data leakage. In such cases, models may learn subject-specific features rather than generalizable patterns, leading to overestimated performance on the test set and poor generalization to new individuals.}

To assess the effect of window-based data overlap on model performance, we designed a synthetic classification task in which the input data instances are divided into sequential windows. In this setup, the first window is used for training and the second for evaluation. We progressively increased the degree of overlap between consecutive windows, simulating a common pitfall in time-series or segment-based tasks where test data may partially appear in the training set. A simple two layer DNN with 100 and 50 neurons respectively was used for evaluation. Performance are averaged across 10 runs on randomly generated synthetic data. Results are presented in Fig. \ref{fig:exp:data_overlap_leakage}.
 We evaluate how this overlap ratio influences the classifier's ability to generalize. Performance is measured as classification accuracy averaged over multiple random experiments.
 
 As expected, the inclusion of overlapping samples (red) in the training set leads to artificially improved accuracy respect to leakage-free condition (blue), emphasizing the need for strict temporal or structural separation between train and test segments.

\begin{figure}
    \centering

\begin{tikzpicture}

\definecolor{darkgray176}{RGB}{176,176,176}

\begin{axis}[
tick align=outside,
tick pos=left,
title={Window Overlap Leakage Effect (with Mean ± Std)},
x grid style={darkgray176},
xlabel={Overlap ratio between windows},
xmajorgrids,
xmin=-0.045, xmax=0.945,
xtick style={color=black},
y grid style={darkgray176},
ylabel={Accuracy (± std)},
ymajorgrids,
ymin=0.447085264503127, ymax=0.977935819140187,
ytick style={color=black}
]
\path [draw=blue, semithick]
(axis cs:0,0.490223611165369)
--(axis cs:0,0.525776388834631);

\path [draw=blue, semithick]
(axis cs:0.1,0.471214835168448)
--(axis cs:0.1,0.554785164831552);

\path [draw=blue, semithick]
(axis cs:0.2,0.520702941459222)
--(axis cs:0.2,0.551297058540778);

\path [draw=blue, semithick]
(axis cs:0.3,0.592289145532408)
--(axis cs:0.3,0.657710854467592);

\path [draw=blue, semithick]
(axis cs:0.4,0.631833687944868)
--(axis cs:0.4,0.710166312055132);

\path [draw=blue, semithick]
(axis cs:0.5,0.692350889359326)
--(axis cs:0.5,0.717649110640673);

\path [draw=blue, semithick]
(axis cs:0.6,0.687853257842072)
--(axis cs:0.6,0.756146742157928);

\path [draw=blue, semithick]
(axis cs:0.7,0.733871663859499)
--(axis cs:0.7,0.804128336140501);

\path [draw=blue, semithick]
(axis cs:0.8,0.802726381504504)
--(axis cs:0.8,0.821273618495496);

\path [draw=blue, semithick]
(axis cs:0.9,0.848702941459222)
--(axis cs:0.9,0.879297058540778);

\addplot [semithick, blue, mark=-, mark size=5, mark options={solid}, only marks]
table {%
0 0.490223611165369
0.1 0.471214835168448
0.2 0.520702941459222
0.3 0.592289145532408
0.4 0.631833687944868
0.5 0.692350889359326
0.6 0.687853257842072
0.7 0.733871663859499
0.8 0.802726381504504
0.9 0.848702941459222
};
\addplot [semithick, blue, mark=-, mark size=5, mark options={solid}, only marks]
table {%
0 0.525776388834631
0.1 0.554785164831552
0.2 0.551297058540778
0.3 0.657710854467592
0.4 0.710166312055132
0.5 0.717649110640673
0.6 0.756146742157928
0.7 0.804128336140501
0.8 0.821273618495496
0.9 0.879297058540778
};
\path [draw=red, semithick]
(axis cs:0,0.490223611165369)
--(axis cs:0,0.525776388834631);

\path [draw=red, semithick]
(axis cs:0.1,0.486861378000815)
--(axis cs:0.1,0.559138621999185);

\path [draw=red, semithick]
(axis cs:0.2,0.542303061543301)
--(axis cs:0.2,0.571696938456699);

\path [draw=red, semithick]
(axis cs:0.3,0.606364357873447)
--(axis cs:0.3,0.663635642126553);

\path [draw=red, semithick]
(axis cs:0.4,0.669014288630928)
--(axis cs:0.4,0.738985711369072);

\path [draw=red, semithick]
(axis cs:0.5,0.746338096210309)
--(axis cs:0.5,0.769661903789691);

\path [draw=red, semithick]
(axis cs:0.6,0.748100251257868)
--(axis cs:0.6,0.787899748742132);

\path [draw=red, semithick]
(axis cs:0.7,0.801808398292582)
--(axis cs:0.7,0.854191601707418);

\path [draw=red, semithick]
(axis cs:0.8,0.862928932188134)
--(axis cs:0.8,0.877071067811865);

\path [draw=red, semithick]
(axis cs:0.9,0.928193751525134)
--(axis cs:0.9,0.953806248474866);

\addplot [semithick, red, mark=-, mark size=5, mark options={solid}, only marks]
table {%
0 0.490223611165369
0.1 0.486861378000815
0.2 0.542303061543301
0.3 0.606364357873447
0.4 0.669014288630928
0.5 0.746338096210309
0.6 0.748100251257868
0.7 0.801808398292582
0.8 0.862928932188134
0.9 0.928193751525134
};
\addplot [semithick, red, mark=-, mark size=5, mark options={solid}, only marks]
table {%
0 0.525776388834631
0.1 0.559138621999185
0.2 0.571696938456699
0.3 0.663635642126553
0.4 0.738985711369072
0.5 0.769661903789691
0.6 0.787899748742132
0.7 0.854191601707418
0.8 0.877071067811865
0.9 0.953806248474866
};
\addplot [semithick, blue, mark=o, mark size=3, mark options={solid,fill opacity=0}, only marks]
table {%
0 0.508
0.1 0.513
0.2 0.536
0.3 0.625
0.4 0.671
0.5 0.705
0.6 0.722
0.7 0.769
0.8 0.812
0.9 0.864
};
\addplot [semithick, red, mark=x, mark size=3, mark options={solid}, only marks]
table {%
0 0.508
0.1 0.523
0.2 0.557
0.3 0.635
0.4 0.704
0.5 0.758
0.6 0.768
0.7 0.828
0.8 0.87
0.9 0.941
};
\end{axis}

\end{tikzpicture}

\caption{\newtexttwo{data-overlap leakage: The x-axis shows the proportion of overlap between the training and test windows, ranging from 0 (fully disjoint) to 1 (full overlap). The y-axis indicates the average classification accuracy. In the clean setting (blue), training and test windows are strictly separated. In the leaky condition (red), a portion of the test window is included in the training data. As the overlap increases, model performance appears inflated due to partial exposure to test data during training. }}
    \label{fig:exp:data_overlap_leakage}
\end{figure}

\color{black}

Furthermore, when dealing with a problem that relies on sequential data, it becomes crucial to consider this aspect during the dataset split to avoid risks of leakage based on temporal sequences. In other words, future data may contain information that enables the model to make accurate inferences about the past, allowing the model to learn "from the future". However, several tasks involving ML models are typically designed to handle recent incoming data, making evaluation on past data useless. \newtexttwo{In Fig. \ref{code:learnfromfuture_leak}, we consider a dataset $D$ composed of instances having a temporal relationship between them (for example, a time series). If the split strategy does not take into account the time relation as in Fig. \ref{code:learnfromfuture_leak}a and the task is to make prediction about the future, $D^E$ can be composed of data instances coming temporally before from data instances in $D^T$. Consequently, the trained model may be influenced by information from the future with respect to the evaluation, leading to potential biases in assessing its predictive capabilities. To address this temporal concern, a split strategy that respects the chronological order of the data can be implemented, as described in Fig. \ref{code:learnfromfuture_leak}b. In other words, data instances in $D^T$ should precede those in $D^E$ in terms of temporal ordering, ensuring that the model is trained on historical data and evaluated on future instances, leading to a more realistic assessment of its predictive performance in scenarios where the task involves making predictions about upcoming observations.}
Ideally, the test set should consist only of the most recent data, leaving the rest for the training set \citep{bergmeir2012use}. However, achieving this ideal scenario might be challenging due to data scarcity.   
\begin{figure}
\input{_IMG_CODE_SEQUENCE_LEAK}
\label{code:learnfromfuture_leak}
\end{figure}
This issue is also addressed in \citep{plotz2021applying}, which discusses consecutive segments of sensor data, and in \citep{lyu2021empirical}, which examines the use of random splits in time-related data for AIOps. 
Moreover, conventional classical splitting strategies involved in evaluation procedures (such as classical $k$-fold Cross Validation \citep{stone1978cross}) often do not account for this situation \citep{park2012flaws}, leading to unreliable evaluations if adopted in these cases. \citep{kapoor2023leakage} identified a study presented in literature affected by this type of data leakage due to the adoption of  $k$-fold with temporal data. This issue is usually attenuated considering evaluation strategies where the test set is composed only from data coming from the final part of the time series. In \citep{bergmeir2012use} an examination of these strategies is conducted, including ad-hoc modified $k$-fold cross-validation schemes tailored for time-series data. However, the study is limited to stationary time series. In the case of non-stationary time series (e.g., EEG data \citep{kamrud2021effects}), the conclusions drawn may differ, emphasizing the need to attention when dealing with different temporal characteristics. Indeed, the violation of the independent and identically distributed (i.i.d.) assumption makes classical cross-validation procedures potentially unsuitable for the given task \citep{arlot2010survey}.
Additionally, when determining the data split strategy, it's essential to consider these aspects to minimize the risk of leakage, such as avoiding the sharing of information resulting from overlaps between data instances across training and test sets. Similar considerations can be made in case of data acquired from sources with different backgrounds about the problem and the experimental assessment (see Sec. \ref{sec:dataleaktypes:inner}). This condition, if not properly managed both during the dataset preparation and the split strategy stages, can lead to potential leakage conditions. 
\newtexttwo{In Fig. \ref{code:distribution_leak} this situation is described. The dataset comprises data collected from $J$ different distributions, each kept separate. In the scenario reported in \ref{code:distribution_leak}a, a conventional approach for creating training $D^T$ and evaluation $D^E$ sets involves pooling all the data together, without considering the specific distribution sources. The only constraint typically used is the absence of intersection between $D^T$ and $D^E$. However, depending on the nature of the task, this constraint may be insufficient to prevent distribution leakage.
The risk lies in the potential bias introduced when peculiarities of the distributions present in data of both $D^T$ and $D^E$ impact the learning process. This bias could lead to a model that outperform on $D^E$ but might not generalize effectively to new instances, as it has unintentionally adapted to specific characteristics of the combined distributions. To address this, alternative split strategies can be explored, such as maintaining the separation of data by distribution during the split as represented in Fig. \ref{code:distribution_leak}b, ensuring a clearer delineation between training and evaluation sets based on the distribution sources, minimizing the risk of distribution leakage and promoting a more accurate assessment of the model's generalization capabilities.}
\begin{figure}
\input{_IMG_CODE_DISTRIBUTION_LEAK}
\label{code:distribution_leak}
\end{figure}

To investigate the impact of distribution leakage, we simulated a multi-source classification scenario in which data points originate from distinct underlying distributions (sources). Each source was modeled as a separate dataset generated using a synthetic classification task, and statistical dissimilarity between sources was controlled via an additive shift applied to the input features.

We varied the shift magnitude from 0.0 (fully overlapping distributions) to 5.0 (very separated distributions) to simulate increasing distributional divergence across sources.

A simple two layer DNN with 100 and 50 neurons respectively was used for evaluation. For each level of distributional shift, two data splitting strategies are compared. In the leakage setting (blue), the split is performed by randomly partitioning the entire dataset, without considering that the data originate from different sources—thus potentially mixing samples from the same distribution across training and test sets. In contrast, the clean setting (red) enforces strict source separation: training and test sets are built from disjoint subsets of sources, simulating a domain generalization scenario where the model must generalize to entirely unseen distributions. This setup allowed us to assess how leakage arising from distributional overlap can lead to artificially optimistic performance estimates—especially as the gap between source distributions increases.

The results, presented in Fig. \ref{fig:exp:distribution_leakage}, show that as the distributional shift between sources increases (red), the performance in the leakage-free condition (blue) degrades significantly, reflecting the model's limited ability to generalize across domains. In contrast, the leaky setup, where training and test samples are drawn from a shared distribution, produces consistently higher and misleadingly stable accuracy, masking the true generalization gap.

\begin{figure}
\centering
\begin{tikzpicture}
\definecolor{darkgray176}{RGB}{176,176,176}
\begin{axis}[
tick align=outside,
tick pos=left,
title={Effect of Distribution Leakage vs Shift Between Domains},
x grid style={darkgray176},
xlabel={Shift between sources (distribution distance)},
xmajorgrids,
xmin=-0.25, xmax=5.25,
xtick style={color=black},
y grid style={darkgray176},
ylabel={Accuracy},
ymajorgrids,
ymin=0.472681602267753, ymax=0.927985267140154,
ytick style={color=black}
]
\path [draw=blue, semithick]
(axis cs:0,0.72271035399041)
--(axis cs:0,0.80728964600959);

\path [draw=blue, semithick]
(axis cs:1,0.644309527005433)
--(axis cs:1,0.821690472994567);

\path [draw=blue, semithick]
(axis cs:2,0.537384211741078)
--(axis cs:2,0.742615788258922);

\path [draw=blue, semithick]
(axis cs:3,0.507013514747624)
--(axis cs:3,0.654986485252376);

\path [draw=blue, semithick]
(axis cs:4,0.494386048397744)
--(axis cs:4,0.597613951602256);

\path [draw=blue, semithick]
(axis cs:5,0.493377223398316)
--(axis cs:5,0.556622776601684);

\addplot [semithick, blue, mark=-, mark size=5, mark options={solid}, only marks]
table {%
0 0.72271035399041
1 0.644309527005433
2 0.537384211741078
3 0.507013514747624
4 0.494386048397744
5 0.493377223398316
};
\addplot [semithick, blue, mark=-, mark size=5, mark options={solid}, only marks]
table {%
0 0.84728964600959
1 0.821690472994567
2 0.742615788258922
3 0.654986485252376
4 0.597613951602256
5 0.556622776601684
};
\path [draw=red, semithick]
(axis cs:0,0.709278754891707)
--(axis cs:0,0.864054578441626);

\path [draw=red, semithick]
(axis cs:1,0.70879773408334)
--(axis cs:1,0.857868932583326);

\path [draw=red, semithick]
(axis cs:2,0.701305199944724)
--(axis cs:2,0.855361466721942);

\path [draw=red, semithick]
(axis cs:3,0.709960328701831)
--(axis cs:3,0.850039671298169);

\path [draw=red, semithick]
(axis cs:4,0.712464868359911)
--(axis cs:4,0.850868464973422);

\path [draw=red, semithick]
(axis cs:5,0.711068388704404)
--(axis cs:5,0.855598277962263);

\addplot [semithick, red, mark=-, mark size=5, mark options={solid}, only marks]
table {%
0 0.709278754891707
1 0.70879773408334
2 0.701305199944724
3 0.709960328701831
4 0.712464868359911
5 0.711068388704404
};
\addplot [semithick, red, mark=-, mark size=5, mark options={solid}, only marks]
table {%
0 0.864054578441626
1 0.857868932583326
2 0.855361466721942
3 0.850039671298169
4 0.850868464973422
5 0.855598277962263
};
\addplot [semithick, blue, mark=o, mark size=3, mark options={solid}, only marks]
table {%
0 0.785
1 0.733
2 0.64
3 0.581
4 0.546
5 0.525
};
\addplot [semithick, red, mark=x, mark size=3, mark options={solid,fill opacity=0}, only marks]
table {%
0 0.786666666666667
1 0.783333333333333
2 0.778333333333333
3 0.78
4 0.781666666666667
5 0.783333333333333
};
\end{axis}

\end{tikzpicture}

\caption{\newtexttwo{The x-axis denotes the degree of shift between the source distributions, simulating increasing domain dissimilarity. Accuracy is reported for two conditions: with distribution leakage (red) and without leakage (blue). As the shift increases, accuracy in the leakage-free condition drops significantly, reflecting the model's difficulty in generalizing across domains. In contrast, the leakage condition where training and test samples are drawn without preserving domain separation inflated performance across all shift levels.}}
\label{fig:exp:distribution_leakage}
\end{figure}

\color{black}

As already discussed in Sec. \ref{sec:background:indvstrans}, in Transductive ML all the available data and information can be exploited during the model training, since the goal is not to obtain a model able to generalize on new data.
Instead, in TL cases, again it is important to consider the task goals. Considering DA methods, in several works in the literature the training data is considered as ${D}_{source}$, while the data used to assess the model (i.e., the test set) is regarded as ${D}_{target}$. It is often not clear if there is a split between training and evaluation target data,  e.g., \citep{zheng2015transfer,zhang2019cross,long2014transfer,sun2016return,chai2016unsupervised,chai2017fast,chai2018multi,luo2018wgan}. However, what should better clarified is the fundamental assumptions regarding the paradigm adopted and the proposed goal. For example, considering the scenarios described in Sec. \ref{sec:dataleaktypes:splitrelated:trans}, the data leakage condition issue described does not occur if the task is intra-subject rather than inter-subject. In intra-subject, both the source and target domains consist of data from the same subject, but acquired at different times. Since all the data belongs to a single subject, the model can effectively leverage the shared patterns across different acquisitions, which are specific to that subject. In this particular case, shared patterns become an advantage because they represent characteristics unique to the target subject, making the model more robust and tailored to the target subject.
More in general, before evaluating the potential for data leakage, it is essential to clarify the goal of the ML task.
If the objective of the DA method is to operate effectively on available target data during the training without a broader goal of generalization, employing all the available data, including those belonging to the evaluation sets, during the training stage can be deemed safe. However, in this case no further generalization assumptions can be made regarding data belonging to the target domain acquired outside the training stage. Indeed, the assessed performance is limited to the target data used during the DA training process, corresponding to the whole test dataset, and cannot be considered valid for other data outside this scope. This type of leakage for DA is summarized in Fig. \ref{code:da_distribution_leak}. \newtexttwo{In particular, let's consider a training function, denoted as $train^{DA}$, which takes as input a model $M$ and two datasets, $D_{source}$ and $D_{target}$, collected from the source and target domains, respectively. If the objective is to transfer knowledge from the domain of $D_1$ to the domain of $D_2$, a potential pitfall arises when training the model $M$ by considering as data originating from the target domain the entire $D_2$.
In scenario described in Fig. \ref{code:da_distribution_leak}a, where the entire $D_2$ dataset is used for training and subsequently evaluating the model, there is a risk of data leakage. The model might inadvertently adapt to the specific characteristics of $D_2$, leading to an optimistic evaluation on the same data. This scenario undermines the model's ability to generalize to new instances beyond the training domain.
To mitigate the risk of distribution leakage, an alternative approach is described in Fig. \ref{code:da_distribution_leak}b. Here, the $D_2$ dataset is split, allocating a portion of the data ($D_2^A$) for the training stage and reserving another portion ($D_2^B$) to evaluate the model. This separation helps ensure that the model is not evaluated on the same data it was trained on, providing a more accurate assessment of its generalization performance to the target domain of $D_2$.}
\begin{figure}
\input{_IMG_CODE_DA_DISTRIBUTION_LEAK}
\label{code:da_distribution_leak}
\end{figure}

Conversely, if the goal is to generalize toward new data becoming to the target domain but not available during the training, using the test data entirely during training is not advisable. Instead, considering DG strategies, we assumes that several datasets from different source domains are available, with no available data from the target domain during training. A potential dataset ${D}_{target}$ from a target domain is solely accessible for evaluating the trained model. Within the DG framework, a clear distinction is established between the data utilized for training (comprising the ${D}_{source_j}$ sets, where $1\leq j \leq S$) and data exclusively used for assessing model performance. Maintaining this division should prevent sets' intersection leakage. On the other side, distribution leakage can happen if the task hasn't been properly defined. 

%
\section{Final discussion and considerations}
\label{sec:final_disc}

Data leakage can significantly impact a model's performance by artificially inflating its evaluation results, undermining its generalization ability to unseen data. 
In this work, it has been shown to identify data leakage is not a naive task. In particular, defining the task and specifying the goals of the experiment are crucial to correctly determining whether data leakage can be present. Furthermore, we  proposed a taxonomy of data leakage scenarios composed of three main families: Preprocessing-related leakage, Split-related leakage, and Data-induced leakage. For each family, we analyze possible data leakage scenarios for classical ML, transductive ML, and TL.

Data-induced leakage arises from flaws during data collection, such as spurious features or artifacts in the data, which persist even after correct splitting. For instance, models might exploit irrelevant patterns that don't generalize to new data. Preventing this requires a clear definition of the task and the careful selection of features. Leakage can also occur when real labels are inferred from input features. In transductive ML data leakage occurs less frequently because, in this setting, the goal is not generalization to unseen data but rather solving the task using all available data. Scenarios that in classical ML could lead to leakage, such as oversampling or feature engineering based on the entire dataset, in a transductive setting are permissible since no additional data beyond the provided dataset is expected. Instead, in TL, if the relationship between source and target domain data is not properly considered during the task definition and the experimental setup, data leakage can easily occur. For example, data collected from different time windows in EEG studies may have hidden correlations that mislead the model into learning spurious patterns, causing inflated performance that doesn’t reflect true generalization. Furthermore, leakage can happen when fine-tuning models on datasets that overlap with the pre-training data, leading to "bias transfer learning" issues where models benefit from previously seen evaluation data.

Preprocessing-related leakage occurs when information from the evaluation data is inadvertently used during preprocessing steps contaminating the training process. Inappropriate application, especially without splitting data beforehand, can result in over-optimistic evaluation. For instance, improper oversampling or undersampling can lead to leakage if the synthetic samples influence evaluation data, skewing performance metrics. A good rule of thumb is to apply sampling techniques only after data splitting. Additionally, sequential or time-series data need specific attention in splitting strategies to avoid leakage due to temporal dependencies. Also in TL scenarios, where the goal may be to generalize to unseen data from a target domain, practices such as normalization or feature engineering can cause leakage. In these cases, target domain data should not be involved in preprocessing to prevent leakage.

Split-related leakage is caused by inappropriate data splitting strategies for the task at hand, such as having overlaps or shared information between the training and test sets.  However, split-related leakage can also propagate effects from earlier data management issues, and appropriate split strategies must align with the task at hand, particularly in domain-specific scenarios. As in other cases, the validity of potential leakage depends on the task. For instance, the setup in \citep{lee2018validity}, which might appear to constitute data leakage under standard assumptions, is appropriate for the intended task, as discussed in Sec. \ref{sec:dataleaktypes:splitrelated:general_disc}.
In TL, however, the task's goals say whether to use all available data to solve the task is allowed. For instance, in DA, using all available data, including target domain data during training, can be acceptable if the goal is not to generalize beyond the target domain. Conversely, if the aim is to generalize to new, unseen data from the target domain, using the test data for training leads to misleadingly high performance. In DG, which aims to generalize without knowledge about the target domain, split-related leakage can occur if data closely related are present in both the training and test sets, leading the model to recognize familiar patterns rather than learn generalizable features. Defining the task, again, is essential to avoid these issues. Summarizing, in transductive ML the concern of data leakage disappears or is mitigated because using all the possible information is part of the learning process. In TL, instead, the distinction between useful transferred knowledge and data leakage becomes more subtle. The goal is to use information from the available data from several domains \textit{to generalize} (therefore as classical inductive ML) in another domains, but without leaking from the evaluation data of the target domain(s), undermining the reliability of the evaluated performance. 
\newtext{In Tab. \ref{table:safe-unsafe-practices} a summary of best practice and common misleading practices is provided.}
\begin{table}[ht!]
\begin{adjustbox}{angle=90}
\scalebox{0.7}{
\begin{tabular}{|m{2cm}|m{2cm}|m{10.5cm}|m{10.5cm}|}
\hline
\centering\arraybackslash\textbf{Type} & \centering\arraybackslash\textbf{Main pipeline steps} & \centering\arraybackslash\textbf{Safe Practices} & \centering\arraybackslash\textbf{Examples of misleading practices} \\ 
\hline
\multicolumn{4}{|c|}{\textbf{Data-induced Leakage} (Sec. \ref{sec:dataleaktypes:inner})}\\
\hline
\centering\arraybackslash\textbf{Data Collecting Leakage} \newline(Sec. \ref{sec:dataleaktypes:inner:collect}) &
Data management (collection) &
- Carefully select data acquisition sources, ensuring any dependencies between training and evaluation data align with the task. \newline
- Design a clear and task-specific data acquisition procedure, minimizing artifacts or dependencies between data instances that could bias the model's evaluation. \newline
- In TL, assess if the data acquired from different domains can influence each other. \newline
- The collected dataset is entirely new and does not include data from any existing datasets.
&
- The data acquisition procedure generates hidden artifacts or dependencies which can affect the model's evaluation for the inspected task. \newline
- The used dataset is composed of other datasets, resulting in duplicated data appearing in both the training and evaluation sets. \newline
- Artifacts or patterns affect the training and evaluations sets due to an experimental setup not properly designed. \newline
- In TL, the pre-trained model may be inadvertently fine-tuned on data related to the original pre-training phase, affecting the final evaluation due to unintended dependencies.\\
\hline
\centering\arraybackslash\textbf{Label Leakage} \newline (Sec. \ref{sec:dataleaktypes:label}) & 
Dataset construction (acquisition, data instance design) &
- Evaluate which features are appropriate to be included based on the specific task, focusing on those that may be indirectly related to the real labels or result from knowledge of the labels themselves. & 
- All available data are used to maximize information, without considering the task and how these features are set. \\
\hline
\centering\arraybackslash\textbf{Synthesis Leakage} \newline (Sec. \ref{sec:dataleaktypes:synthesis}) & 
Dataset construction (instances synthesis, data analysis) &
- Perform oversampling/undersampling only after splitting into training and evaluation sets. \newline
- In TL, ensure that data from different domains does not share artifacts or patterns not related to the task. & 
- Perform oversampling/undersampling before splitting into training and evaluation sets to maximize information.\\
\hline
\multicolumn{4}{|c|}{\textbf{Preprocessing Leakage} (Sec. \ref{sec:dataleaktypes:preproc}) }\\
\hline
\centering\arraybackslash\textbf{Normalization, Imputation, Cleaning, Feature Engineering leakage}\newline (Sec. \ref{sec:dataleaktypes:preproc:norm}-
\ref{sec:dataleaktypes:preproc:feat}) & 
Data split, Preprocessing &
- Compute preprocessing parameters functions (e.g., normalization, imputation) only on the training data. &
- Applying preprocessing, together with their parameters' computation, across the whole dataset before splitting to maximize information. \newline
- In TL, compute preprocessing parameters by also leveraging evaluation data from the target domain.\\
\hline
\multicolumn{4}{|c|}{\textbf{Split-related Leakage} (Sec. \ref{sec:dataleaktypes:splitrelated})}\\
\hline
\centering\arraybackslash\textbf{Similarity leakage (Sets' Intersection)}\newline (Sec. \ref{sec:dataleaktypes:setsintersect}) & 
Dataset construction, Data split, model training &
- Ensure that the evaluation set reflects the structural characteristics of future samples. If the task does not involve identical data instances, ensure that no data instances are shared between the training and evaluation datasets.&
- Allowing the training and evaluation sets to share the same instances, also during the training stage when an evaluation step is involved (e.g., early stopping).\newline \\
\hline
\centering\arraybackslash\textbf{Similarity leakage (Overlap)}\newline  (Sec. \ref{sec:dataleaktypes:similarity}) & 
Dataset construction (data instance design), data split &
- Ensure that the data instance design, along with the chosen split strategy, does not result in shared information between training and evaluation sets. & 
- Data instances in the training and evaluation sets share features due to the adopted data instance design and split strategy (e.g., splitting temporal sequences with overlapping segments).\\
\hline
\centering\arraybackslash\textbf{Distribution Leakage} \newline (Sec. \ref{sec:dataleaktypes:distribution}) & 
dataset construction, data split, model training &
- Ensure splits are based on realistic task-related distributions.\newline
- Consider whether the task is focused on capturing population-specific characteristics of the data or aims for broader generalization across different populations.\newline
- In case the task involves temporal relations, ensure temporal splits respect real-world data flow (past vs future). In particular, only past data should be used for training, while future data should remain unseen.\newline 
- In TL, reserve a portion of target domain data for evaluation.\newline
&
- In TL, evaluating the model on the whole target dataset, even though it was used in the training stage. \newline
- Randomly splitting data without considering real-world distribution or time-based relationships (e.g., mixing past and future data in training and evaluation). \newline
- Treating data from different sources as if they belong to the same source, when the task aims to generalize across different sources. \\
\hline
\end{tabular}
}
\end{adjustbox}
\caption{Safe practices to avoid common data leakage cases, with the main pipeline steps where leakage can occur, and examples of misleading practices.}
\label{table:safe-unsafe-practices}
\end{table}

\clearpage

\color{black}
\section{Conclusion}
\label{sec:conclusion}
In this paper, we investigated the complexities surrounding data leakage in the context of ML. As ML continues to be part of several research domains, it is important to address the unintended consequences of data leakage, especially in scenarios where users and practitioners adopt a ``black box'' approach, relying on simplified ML tools without a deep understanding of the underlying ML methodologies. While this approach democratizes access to ML capabilities, it inadvertently introduces challenges, with data leakage emerging as a central concern. The lack of meticulous consideration in task addressed, paradigm adopted, pipeline workflow, and model evaluation protocol can lead to the unintentional inclusion of forbidden information during model training, compromising performance evaluation. Therefore, while the ``push the button'' approach facilitates accessibility to ML, it's essential for practitioners to comprehend the fundamentals and potential implications to make informed decisions and ensure optimal utilization of ML in their applications.
\newtexttwo{We emphasized the connection between data leakage occurrences and the specific tasks being addressed. Indeed, understanding the interplay between task and data leakage is crucial for mitigating risks and ensuring the robustness of ML applications.
Moreover, our discussion extended to unconventional ML paradigms, such as transductive ML, where the distinctions between inductive and transductive approaches, if overlooked, can exacerbate data leakage issues. Furthermore, data from different domains in TL frameworks can introduce unintended information, challenging the traditional assumptions of model generalization. Therefore, we delineated instances where data leakage may or may not occur, providing  insights for practitioners.}

In conclusion, this paper discussed about data leakage propagation, task-dependent nature, and relevance in TL. We aim to empower practitioners and researchers to adopt a more informed and responsible approach to the utilization of ML technologies in diverse domains.
By incorporating task-specific knowledge and adopting proper mitigation strategies, the evaluation stage can be fortified against unintended influences, ensuring a more reliable assessment of ML models.

\section*{Funding}
This work was partially supported by the {Ministry of University and Research, PRIN research project ``BRIO -- BIAS, RISK, OPACITY in AI: design, verification and development of Trustworthy AI.''}, Project no. 2020SSKZ7R.

\newpage
\bibliography{_bib}
\begin{appendices}

\end{appendices}

\end{document}